\documentclass[pdflatex,sn-mathphys-num]{sn-jnl}

\usepackage{placeins}
\usepackage{graphicx}%
\usepackage{multirow}%
\usepackage{amsmath,amssymb,amsfonts}%
\usepackage{amsthm}%
\usepackage{mathrsfs}%
\usepackage[title]{appendix}%
\usepackage{xcolor}%
\usepackage{textcomp}%
\usepackage{manyfoot}%
\usepackage{booktabs}%
\usepackage{algorithm}%
\usepackage{algorithmicx}%
\usepackage{algpseudocode}%
\usepackage{listings}%
\usepackage{booktabs}
\usepackage{makecell}
\usepackage{subcaption}
\usepackage{url}
\usepackage{tabularx}
\usepackage{hyperref}
\usepackage{geometry}
\usepackage{lmodern}
\usepackage{anyfontsize}
\usepackage[normalem]{ulem} 

\begin{document}

\title[Domain Adaptation-Enhanced Searchlight]{Domain Adaptation-Enhanced Searchlight: Enabling classification of brain states from visual perception to mental imagery}


\author*[1]{\fnm{Alexander} \sur{Olza}}\email{alexander.olza@ehu.eus}

\author[2,3]{\fnm{David} \sur{Soto}}\email{d.soto@bcbl.eu}

\author[1]{\fnm{Roberto} \sur{Santana}}\email{roberto.santana@ehu.eus}

\affil*[1]{\orgdiv{Intelligent Systems Group}, \orgname{University of the Basque Country (UPV/EHU)}, \orgaddress{\city{Donostia - San Sebastián}, \country{Spain}}}

\affil[2]{\orgdiv{Consciousness Group}, \orgname{Basque Center for Cognition, Brain and Language (BCBL)}, \orgaddress{\city{Donostia - San Sebastián}, \country{Spain}}}

\affil[3]{\orgname{Ikerbasque, Basque Foundation for Science}, \orgaddress{\city{Bilbao}, \country{Spain}}}


\abstract{In cognitive neuroscience and brain-computer interface research, accurately predicting imagined stimuli is crucial. This study investigates the effectiveness of Domain Adaptation (DA) in enhancing imagery prediction using primarily visual data from fMRI scans of 18 subjects. Initially, we train a baseline model on visual stimuli to predict imagined stimuli, utilizing data from 14 brain regions. We then develop several models to improve imagery prediction, comparing different DA methods. Our results demonstrate that DA significantly enhances imagery prediction in binary classification on our dataset, as well as in multiclass classification on a publicly available dataset. We then conduct a DA-enhanced searchlight analysis, followed by permutation-based statistical tests to identify brain regions where imagery decoding is consistently above chance across subjects. Our DA-enhanced searchlight predicts imagery contents in a highly distributed set of brain regions, including the visual cortex and the frontoparietal cortex, thereby outperforming standard cross-domain classification methods. The complete code and data for this paper have been made openly available for the use of the scientific community. }

\keywords{Domain Adaptation, Brain Decoding, fMRI, Searchlight}



\maketitle

\section{Introduction}\label{sec1}

The mapping of brain activation patterns to their corresponding cognitive states is known as brain decoding. Methods for brain decoding are instrumental in understanding the brain processes involved in cognition \cite{elston2022decoding} and therefore an increasing number of works address the question of solving brain decoding tasks with machine learning algorithms \cite{tu2018novel,gupta2021obtaining,saeidi2021neural}. While ``brain decoding'' can encompass various tasks, including visual stimulus reconstruction, this study specifically focuses on the classification of neural patterns associated with visual perception and mental imagery.\\

Specifically, the degree to which visual imagery depends on the same neural mechanisms as visual perception remains a subject of ongoing research. Previous studies using standard multivoxel pattern classification analyses (MVPA) showed that a classifier trained to discriminate the contents of perception in the visual cortex can predict the contents of visual imagery \cite{o2000mental,reddy2010reading,johnson2014decoding,naselaris2015voxel}. However, evidence that perception and imagery share neural representation in frontoparietal areas is scarce \cite{dijkstra2017vividness,ragni2020decoding}.\\

In the present study, we hypothesize that failures to reveal similar neural representations may be due to the classification pipeline not accounting for the different distributions of the data across perception and imagery domains. Consequently, we claim that Domain Adaptation (DA) methods \cite{daume2006domain,zhang2013domain} are an adequate approach to alleviate the hypothesized distribution shift and enhance the transferability of classifiers trained in visual perception to imagery testing data.\\

Thus, we investigate the impact of different DA approaches in a particular fMRI cross-domain experiment where subjects were asked to solve perceptual and imagery tasks. Cross-domain classification tasks involving imagery are the basis for Brain-Computer Interfaces (BCIs) \cite{li2019decoding} and have been addressed before \cite{reddy2010reading,cichy2012imagery,johnson2014decoding,soto2020decoding} but, to our knowledge, the literature is scarce when it comes to applying DA techniques to such problems using fMRI data. Furthermore, rather than restricting our analysis to the visual cortex, we explore the whole brain using one local neighbourhood classifier per voxel (a searchlight procedure, as introduced by \cite{kriegeskorte2006information}). In this procedure, the brain is partitioned into thousands of overlapping spheres and the voxels of each sphere are assigned to a classifier. The searchlight method combats the curse of dimensionality by limiting the radius of the sphere, and offers an anatomically interpretable output.\\

In this paper, we integrate a DA technique into the searchlight method to unveil common neural representations for perception and imagery throughout the brain.\\

In summary, the contributions made in this paper are as follows. 
\begin{enumerate}
    \item First extensive evaluation and comparison of well-established DA techniques on cross-domain classification problems defined on fMRI data, considering several brain Regions Of Interest (ROIs) and two independent datasets. 
    \item Introducing the DA-enhanced searchlight approach using two different DA techniques on the whole brain, to gain more understanding on the nature and location of the domain-shift.
  
\end{enumerate}

\section{Methodological background}
We define a domain $D$ as a combination of input and output spaces and an associated probability distribution: $(X,Y,p_D)$. In the simplest case, there is a source domain $S$ with enough data to build a reliable learner, and our aim is to generalize to a different target domain $T$ for which there is limited knowledge. In this paper, we consider the situation in which both domains share input and output spaces, meaning that they have the same dimensionality and live in the same space, but data come from different probability distributions $p_S$ and $p_T$ .\\

From the ML point of view, the problem we are interested in is risk minimization in the target domain, given a set of data $(x_i,y_i)\sim p_S(x,y)$ from the source distribution (following notation from \cite{kouw2018introduction}). Although the formulation is valid for a general output space $Y$, we will restrict ourselves to binary classification. In this setting, we would like to find the best learner $h^\star$, defined as the one with minimum expected loss $l$ under the target distribution, $h^\star = arg\underset{h\in \mathcal{H}}{min} R_T(h)$, where $h$ is any learner included in a given hypothesis space $\mathcal{H}$. The expression for the risk includes a term $p_T(x_i,y_i)$, that represents the probability of the source samples under the target distribution (Eq. \ref{eq:risk}, left-hand side).\\

Introducing into the integral the probability of the observations coming from their actual source distribution $p_S$, we arrive to an expectation under $p_S$ of the loss multiplied by the ratio of target to source distributions (Eq. \ref{eq:risk} right-hand side).

\begin{equation}\label{eq:risk}
     R_T(h)=\sum_{y\in Y}\int_X l(h(x)|y)p_T(x,y)dx = \sum_{y\in Y}\int_X l(h(x)|y)\frac{p_T(x,y)}{p_S(x,y)}p_S(x,y)dx 
\end{equation}

In practice, this expectation can be estimated using the sample average (empirical risk minimization) \cite{vapnik1991principles}. Depending on the structure of $p_S$ and $p_T$, the quantity $p_T(x,y)/p_S(x,y)$ can be further simplified by factoring into conditional and marginal distributions, giving rise to special cases of distribution shifts. For example, if the prior probabilities of $y$ differ between domains, but the conditionals $p(x|y)$ are close enough, we have a case of prior shift. Other particular settings include covariate shift (similar posterior distributions $p(y|x)$ for $T$ and $S$) and concept shift (similar data distributions $p(x)$ but $p_T(y|x)\neq p_S(y|x)$). In general, though, the data-set shifts have a more complicated and unknown structure. Depending on the nature of the distribution shift, certain DA approaches may be more suitable than others \cite{farahani2021brief}.\\

 DA requires some previous knowledge of the target domain, be it in the form of some labelled (supervised DA, also known as few-shot learning) or unlabelled (unsupervised DA) instances from the target distribution. The amount of target domain knowledge incorporated into a DA technique is usually too scarce to build a model that relies on that information alone. In fact, for most DA applications, acquiring (or labelling) data from the target domain is difficult and/or expensive, but there is enough source domain data to build and validate a model with an adequate performance of unseen samples from that same source domain.\\

Given that such a model is achieved, generalization to novel examples from the target distribution can be tackled by several strategies, namely (a) instance-based, (b) feature-based and (c) parameter-based approaches, as summarized in Table \ref{tab:DAapproaches}. The naïve inclusion of the available target domain knowledge as additional training samples can also be considered as a rudimentary form of DA.\\

\begin{table}[h!]
\begin{tabularx}{\textwidth}{|>{\setlength\hsize{1.4\hsize}\setlength\linewidth{\hsize}}X|>{\setlength\hsize{.9\hsize}\setlength\linewidth{\hsize}}X|>{\setlength\hsize{.7\hsize}\setlength\linewidth{\hsize}}X|}
\hline
\multicolumn{3}{|c|}{Main approaches to Domain Adaptation}\\ \hline
Methods and key ideas & \multicolumn{2}{|c|}{Examples}\\ \hline
  & Supervised & Unsupervised \\
\hline
1. Instance-based
\begin{itemize}
   \item  Measure the difference between each source domain  and 
target domain instances.
     \item Adjust the domain instance contributions during training.
\end{itemize} &
\begin{itemize}
\item Balanced Weighting \cite{wu2004improving}
\item TrAdaBoost \cite{dai2007boosting}
\item IWN \cite{de2022fast}
\end{itemize} &
\begin{itemize}
\item KMM  \cite{huang2006correcting}
\item ULSIF \cite{kanamori2009least}
\item RULSIF \cite{yamada2013relative}
\end{itemize}\\
\hline
2. Feature-based
\begin{itemize}
\item Look for a space of common  features  with respect to the task on source and target domain.
\item Build a representation under  which $p_S$ and $p_T$ are similar.
\end{itemize} &
\vphantom{2. Feature-based}
\begin{itemize}
\item FA and PRED \cite{daume2009frustratingly}
\item Fine-tuning \cite{finetuning}
\end{itemize} &
\begin{itemize}
\item SA \cite{fernando2013unsupervised} 
\item MCD \cite{MCD}
\item DANN \cite{DANN}
\item Deep CORAL \cite{deepcoral}
\end{itemize}\\
\hline
3. Parameter-based
\begin{itemize}
\item Adapt the parameters of a  pre-trained source-only model to build a suitable model with the same  structure on the target domain.
\end{itemize} &
\begin{itemize}
\item Regular Transfer \cite{chelba2006adaptation}
\end{itemize} &
\\
\hline
\end{tabularx}
\caption{Summary of main approaches to Domain Adaptation}
\label{tab:DAapproaches}
\end{table}

\section{Related Work}

In this section, we review a number of works related to our research. We cover: 1) ML techniques for investigating commonalities in brain mechanisms for visual and imagery tasks;  2) DA methods applied to other neuroscience scenarios. \\

\subsection{ML techniques for investigating commonalities in brain mechanisms for visual and imagery tasks}

There have been numerous attempts to elucidate whether visual perception and imagery arise from similar brain mechanisms. Since the development of brain imaging, the most prominent ML methods used to study this problem are multivariate pattern analysis (MVPA) and deep learning approaches. \\

MVPA considers patterns of activity across multiple brain regions simultaneously, as opposed to univariate analysis, which looks at each voxel independently. It is a widespread method for testing cognitive theories \cite{peelen2023testing}. \\

When it comes to the relatedness of perception and imagery, all published work in the decade of 2010 involves attempts of pure transfer using MVPA \cite{reddy2010reading,cichy2012imagery,akama2012decoding,johnson2014decoding}. These pioneering works evaluate a source-only classifier in the data from the target domain without implementing DA techniques. As we have already discussed, pure transfer is not guaranteed to work, due to a distribution shift between the brain patterns induced by different experimental conditions. Indeed, all the aforementioned papers suffer from low generalization, showing a notable decrease between classifier performance in perception data and the attempted classification of imagery data.\\

With a slightly different approach, \cite{naselaris2015voxel} trained a voxel-wise encoding on visual low level features and then applied it to imagery brain patterns recorded while the participants were remembering a previously memorized work of art. Their model was able to identify the target artwork among a large dataset of options, supporting the idea that low level features are similar between perception and imagery. In 2017, \cite{horikawa2017generic} used several feature extraction methods trained on visual fMRI data to construct a generic decoder for both vision and imagery. They attained around 90\% accuracy for seen object and around 60\% accuracy for imagined object identification, revealing a significant performance gap. A review \cite{dijkstra2019shared} from 2019 concluded that visual imagery and perception share a variety of neural mechanisms in the visual, parietal, and frontal cortex, but that those similarities are modulated by imagery vividness, and that temporal dynamics are different (as studied in 2020, see \cite{xie2020visual}). \\

In 2020 \cite{ragni2020decoding}, MVPA was used to decode stimulus identity in visual imagery of simple shapes and letters, identifying the regions where such decoding was possible. A searchlight analysis revealed cross-domain decoding in the occipital, parietal and inferotemporal cortices.\\

From 2023 onwards, deep learning methods have been employed to reconstruct mental images from brain activity patterns \cite{koide2024mental,ahmadieh2024visual}, study visual object representation with a high spatiotemporal resolution \cite{watanabe2023multimodal} or learn robust representations from EEG recordings \cite{singh2024learning}. In \cite{lee2023neural}, the authors conduct a representational similarity analysis comparing the parameters of a CNN trained to recognize digits and a fMRI data-set containing brain patterns linked to digit recognition. \\

However, as pointed out by the authors in \cite{dijkstra2019shared}, the fact that it is possible to have imagery deficits (e.g. aphantasia) without associated perceptual issues indicates that the simulation processes underlying perception and imagery are not exactly the same. More evidence of these differences was found in \cite{breedlove2020generative}, where hierarchical generative networks reveal distinct features for visual perception and imagery in low-level visual areas.\\

In this work, we claim that DA techniques are an additional tool to investigate and quantify the degree of difference between brain mechanisms associated with perception and imagery, as well as to mitigate the cross-domain brain decoding performance gap shown in \cite{cichy2012imagery, reddy2010reading, margolles2024unconscious}. Firstly, we consider several ROIs to determine the best DA approach for our use case. Then, by conducting a searchlight analysis across the whole brain, we investigate which brain regions differ the most between both domains, and improve the classifier's performance on a local level. The study of a large number of subjects offers some insight into the interpersonal variability of the hypothesized domain-shift, as well as the possibility to discover brain regions where such hypothesized domain-shift occurs on a consistent basis.

\subsection{Other applications of DA in neuroscience}

DA has been used in other neuroscientific settings, mostly in cross-device experiments for medical diagnosis. For instance, in \cite{wang2019identifying}, the authors propose a low-rank DA method for autism diagnosis, aiming to determine a common lower-dimensional representation for data from multiple imaging devices. A very different approach involves the field of optimal transport, in which DA has been used to align the distributions of fMRI images from multiple devices with the goal of brain dementia identification \cite{guan2021multi}. 
Both of these methods fit into the category of feature-based DA, because they entail the projection of source and target domains to a new feature space where the domain shift has been minimized. \\

The use of instance-based approaches has also been reported in the literature. For example, in \cite{wachinger2016domain}, the authors propose a re-weighting scheme for the diagnosis of Alzheimer's disease. In the field of brain tumour segmentation, instance-based DA has been employed to correct sample selection bias amongst MRI image patches \cite{goetz2015dalsa}. In \cite{van2014transfer}, the authors compare the performance of four DA techniques in a cross-device MRI image segmentation task using support vector machines. Specifically, they review three instance-based approaches  (including TrAdaboost \cite{dai2007boosting}) and a parameter-based technique, similar to Regular Transfer \cite{chelba2006adaptation}. \\

Regarding other parameter-based methods, most published works deal with fine-tuning of deep learning models. For example, authors of \cite{ghafoorian2017transfer} study the benefits of fine-tuning in a brain lesion segmentation task, adapting CNN models pre-trained on brain MRI scans. Some other works leverage state-of-the-art computer vision models trained in large-scale multipurpose data-sets, adapting the final layers to their task \cite{swati2019brain, khan2019transfer}.  Beyond neuroscience, more examples of DA in medical imaging are referenced in \cite{guan2021domain}.\\

A recent survey \cite{orouji2024domain} advocates for including DA in the computational neuroscientist's toolkit, highlighting the need to test state-of-the-art DA methods in scenarios of complex data such as neuroimaging.\\ 

The application of ML models to diverse human subjects beyond the training population is an important extension of DA in neuroscience. Cross-subject scenarios introduce additional complexity due to anatomical and functional variability between individual brains. Specifically, anatomical differences mean that each brain may have a distinct number of voxels, representing different feature spaces. Therefore, constructing a common feature space is essential before addressing the ML task. Traditional approaches for anatomical alignment across subjects rely on geometrical transformations, such as Tailarach coordinates, MNI space or hyperalignment techniques \cite{haxbyHyperalignment}. However, these methods have notable limitations, as they focus solely on anatomical alignment and do not account for functional variability in brain activity.\\

Recent studies have framed cross-subject learning as a multi-source DA problem, where each subject constitutes a separate source domain. Most advancements have targeted electroencephalogram (EEG) data, particularly for emotion recognition tasks \cite{bao2021twolevel,liang2021multisource}. For example, in \cite{bao2021twolevel}, the authors propose a two-level DA neural network (TDANN) preceded by a feature generator that maps each subject's data into a unified space. Similarly, in \cite{liang2021multisource}, data from all subjects are first mapped into a shared feature space. Subsequent feature extractors are then used to capture subject-specific representations, and these representations’ joint distributions are aligned in a pairwise manner. However, advanced cross-subject DA methods for fMRI data remain underdeveloped. The high spatial resolution of fMRI data and the fact that each individual brain is different poses significant challenges to perform cross-subject DA. Additional work in this area would benefit from novel developments in hyperalignment algorithms \cite{haxbyHyperalignment} to use DA in a common high-dimensional voxel space across subjects.\\

In summary, many neuroscience questions have benefited from DA, but there is a literature gap when it comes to cross-domain classification. Additionally, most published works in the intersection of neuroscience and DA are focused on developing new approaches for very specific tasks and using only one or a reduced set of DA approaches, without extensive evaluation of standard DA methods. In contrast, we conduct a thorough comparison of existing DA methods for a cross-domain task in a large number of independent data-sets, with the goal of quantifying the domain-shift between visual perception and imagery. 

\FloatBarrier
 
\section{Description of the cross-domain analysis}

\begin{figure}[h]
    \centering
    \includegraphics[width=\textwidth]{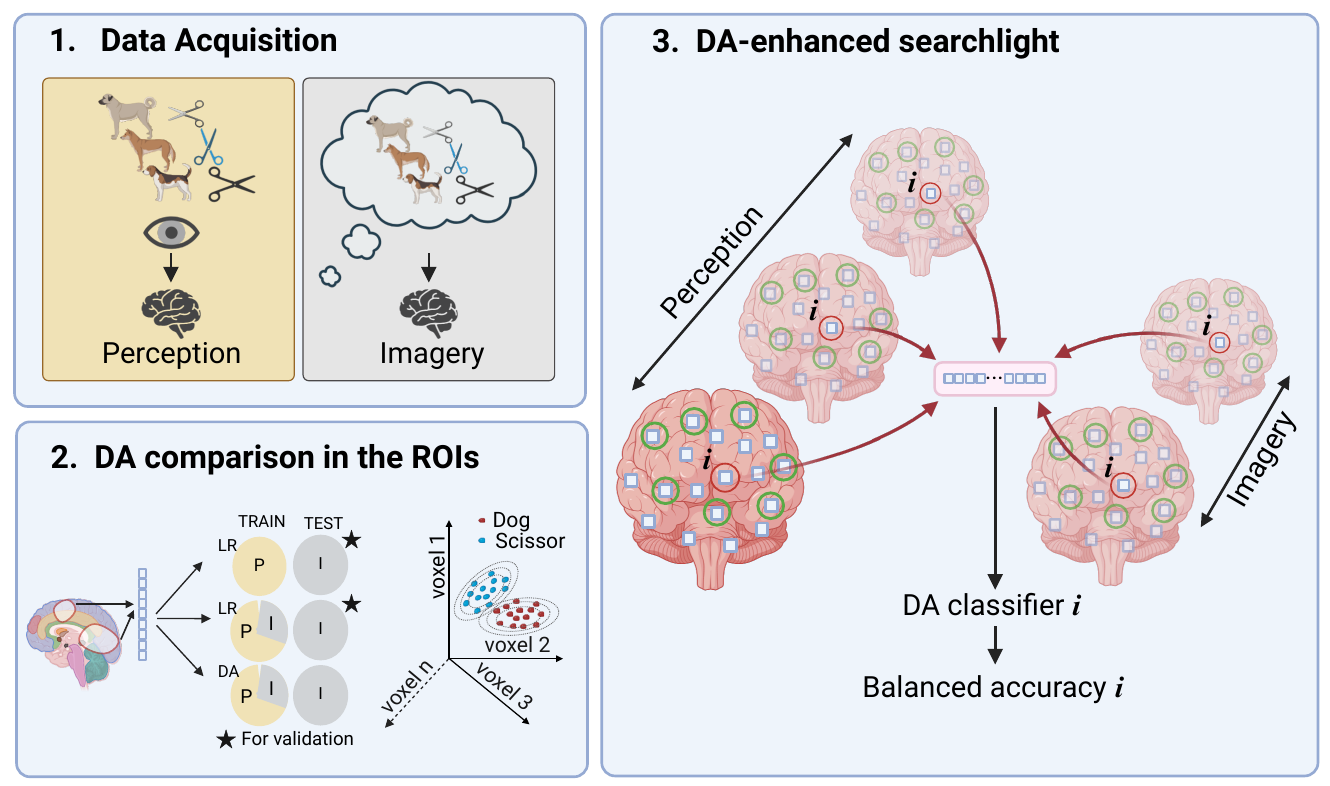}
    \caption{Diagram of the experimental pipeline. Data acquisition: In the perception phase, fMRI data from 18 healthy subjects was acquired while they were presented with pictures of Living (Dog) and Non-living (Scissor) items. In the imagery phase, fMRI scans were recorded while the participants vividly imagined items from each category. DA comparison in the ROIs: The voxels from 14 anatomical Regions of Interest (ROIs) were used to compare 15 DA methods against a Logistic Regression (LR) baseline trained solely on perception and an alternative LR trained on perception and imagery. All the methods were tested on imagery. The comparison was repeated on a publicly available multiclass dataset. DA-enhanced searchlight: A local DA classifier was trained on the vicinity of each voxel in the brain, using data from perception and imagery, to produce a three-dimensional map of classifier accuracy. Created in BioRender. O, A. (2025) \href{https://BioRender.com/u30c785}{https://BioRender.com/u30c785}}
    \label{fig:pipeline}
\end{figure}

Figure \ref{fig:pipeline} summarizes the experimental pipeline detailed in the following subsections.

\FloatBarrier

\subsection{Experimental setup} 

We quantified the contribution of 15 different DA algorithms to a binary cognitive state classification task using 18 data-sets, each containing around 9000-14000 anatomically selected voxels from approximately 500-600 fMRI images of a specific subject gathered in independent realizations of the same experiment, described in detail in \cite{margolles2024unconscious}. The voxels come from 14  ROIs, the exact dimensionality of which is subject-dependent (Supplementary Table \ref{tab:appdescriptive}). The ROIs were extracted using FreeSurfer \cite{Freesurfer}, and their names are listed in Supplementary Table \ref{tab:rois}.\\

The experiment from which the data-sets were collected consisted of two phases: (1) perception and (2) imagery. In the perception step, each participant was presented with living (Dog) and non-living (Scissor) images from the THINGS \cite{THINGS} database. In the imagery stage, they received auditory cues instructing them to vividly imagine an object from one of those two categories for several seconds. The goal was to classify imagery instances into living or non-living, leveraging a classifier trained mostly on perception instances.\\ 

In the perception phase, subjects participated in 4 runs of scanning. Each run was composed of 14 trials, 7 for each category (living/non-living). In each trial, participants observed a sequence of 12 images belonging to the same category. There was an 8.5-second rest period between trials, to ensure separation between brain activity patterns associated with each category. The imagery phase followed an analogous structure but consisted of only 2 runs. \\

Data preprocessing is thoroughly described in \cite{margolles2024unconscious}. The first experimental volume, obtained after the heat-up phase, served as a functional reference. The rest of the volumes  of each participant were co-registered to their corresponding reference volume. To minimize drift, the fMRI time-series was linearly detrended for each run. To capture peak neural activity in response to the presented stimuli, fMRI volumes from 5 to 18 seconds following the stimuli onset were retained (7 volumes per trial). The chosen samples were then independently stacked for region of interest (ROI) delineation. Next, the stacked perceptual volumes were z-scored by subtracting each voxel's mean value and dividing by its standard deviation. Standardization of the imagery run scans was performed using the mean and standard deviation of the perceptual voxels to maintain a consistent and comparable scale across both conditions, facilitating analysis of the generalizability of information from perception to mental imagery.

We also performed the same analysis in a multiclass setting using the publicly available dataset ``Generic Object Decoding'' \cite{horikawa2017generic}, that contains fMRI scans from 5 subjects in both the visual perception and imagery domains. The dataset description and the obtained results are shown in \ref{sec:appendixGOD}.

\subsection{Machine learning task and validation framework}

Taking the delay of the hemodynamic response and the timing of stimulus presentation into account, there was not a one-to-one correspondence between stimuli and fMRI images within each trial, but those from different trials were safely considered independent. Thus, in all our analysis, we avoided data leakage ensuring that all instances of the same trial belonged either to the train or the test set.\\

We used 100 different partitions of the data to validate our methodology. Each partition determined the exact assignment of instances into train or test sets. 
The source domain (perception) train/test ratio was 4:1, with approximate class stratification and observing the trial constraint. As for the target domain, we randomly selected $N_t$ instances with approximate class stratification to compose the training set, discarded the rest of the instances belonging to those trials, and used the remainder for testing. $N_t$ is a parameter of the methodology, which we varied from 10 to 100 in steps of 10.\\

For each train/test split, we fitted a baseline estimator $h$ on the source training set, as well as another instance of the same standard classifier on the union of source and target training sets ($h_{NAIVE}$). We evaluated their performance on the target test set. $h$ is considered to be the experiment baseline, and $h_{NAIVE}$ accounts for the effect of naïvely including some limited knowledge of the target domain into a standard classifier without considering DA techniques. To assess the contribution of a specific DA technique, we fitted a new instance of the chosen estimator on the source domain training set $(X_s, y_s)$ alone and incorporated the target training instances $(X_t, y_t)$ through said DA technique, obtaining a new estimator, $h_{DA}$, which we evaluated on the test sets of the target domain. We chose Logistic Regression as an estimator due to its simplicity, computational speed and compatibility with all DA methods under study. \\

For each subject, we compared the performance of 15 DA approaches on the union of all ROIs using the implementations in the ADAPT \cite{ADAPT} library. \ref{sec:appendix} contains a brief description of each method. Due to the size of the dataset (with around 10 000 features), the Deep Learning methods (FT, DANN, MCD and DCORAL) required a prior dimensionality reduction step. Thus, we performed Independent Component Analysis (ICA) retaining 100 components.\\

As mentioned above, in the first place, we evaluated the contribution of each DA technique to our task in the different subjects using the voxels from all 14 ROIs. Then, based on the comparison results, we also evaluated the performance of the best technique beyond the 14 ROIs, considering all the voxels from the whole brain. This whole-brain analysis was conducted using the searchlight \cite{kriegeskorte2006information} procedure. To evaluate the robustness of the neuroscientific implications of our findings, we repeated this procedure using an alternative DA method.\\

\subsection{Evaluation metrics and statistical validation}

In this section, we detail the validation strategies, both for the initial comparison of DA techniques at the ROI level and for the DA-enhanced searchlight.\\

\subsubsection{Initial comparison of DA techniques}
The evaluation criteria on which our comparisons are based are detailed hereafter. First, note that the methodology was constructed in such a way that the training of $h$, $h_{NAIVE}$ and $\{h_{DA}\}_{DA=1}^{15}$ were performed on exactly the same data, and the same is true for the evaluation phase of all estimators. During the 100 repetitions of the process, the balanced accuracies of each of the 17 estimators were measured on the test set of the target domain, obtaining the 17 corresponding performance vectors of size 100. Those 17 balanced accuracy vectors of $h$, $h_{NAIVE}$ and $\{h_{DA}\}_{DA=1}^{15}$ can be thought of as paired observations. \\ 

To quantify the amount of information needed to actually profit from DA methods, we applied our methodology with a varying amount of target domain instances $N_t$ between 10 and 100 in steps of 10, conducting the same comparison. By doing that, we determined the effect of incorporating a growing amount of target training instances into both $h_{NAIVE}$ and $h_{DA}$; that is, whether $h_{DA}$ profits from this information more than $h_{NAIVE}$ does. \\

Hence, our results are in the form of 10 balanced accuracy tables per subject, one corresponding to each value of $N_t$. Each table has 17 columns (one per ML approach) and 100 rows (one per data partition).\\

Firstly, we concatenated the 10 tables pertaining to each subject and conducted an all versus all statistical comparison based on the Friedman aligned ranks test, with a Shaffer correction for multiple comparisons. The non-parametric character of the Friedman test avoids distributional assumptions on our results, and the Shaffer correction controls the family-wise error rate with a low computational cost compared to other correction methods.\\

We also concatenated all the results (i.e. each of the 180 tables pertaining to each subject and value of $N_t$) to obtain a global comparison between all methods.\\

All the statistical analyses detailed in this section were based on the implementations in the \verb|R| library \verb|scmamp| \cite{scmamp}.\\

\subsubsection{Evaluation of the searchlight analysis}

For the searchlight analysis, we used the same data partitioning methodology as for the ROI-based analysis. In this case, each sample contains a whole-brain three-dimensional image with around 40,000 voxels (the exact number of which is subject-dependent -  see Supplementary Table \ref{tab:appdescriptive}). Each voxel is a cube with a 3mm side, and we defined the radius of the searchlight spheres to be 12mm. We used algorithms RTLC and BW (see \ref{sec:appendixBW} for the results of the latter). \\

We fitted three independent Logistic Regression classifiers on each sphere:

\begin{enumerate}
    \item $h$: Trained only on the samples from the source domain to establish a baseline.
    \item $h_{NAIVE}$: Trained on the union of samples from the source domain and 100 samples from the target domain, to assess the effect of including imagery data without DA.
    \item $h_{DA}$: Incorporating the same 100 samples from the target domain into $h$ via the RTLC or BW algorithm.
\end{enumerate}

All classifiers were tested on the same holdout instances from the target domain. $h$ was also tested on a hold-out set from the source domain. The procedure was repeated for 100 different data partitions, and the mean balanced accuracy per sphere was computed, obtaining a three-dimensional brain map of performance scores for each subject and classifier. To carry out these analyses, we integrated some functionality from \verb|ADAPT| into the searchlight tools from the \verb|Nilearn| library \cite{Nilearn}.\\ 

Then, we subtracted 0.5 (the chance level balanced accuracy) from all the maps for further statistical testing.\\

To be able to extract meaningful anatomical insights and make comparisons between subjects, each score map was transformed to the standard Montreal Neurological Institute (MNI) space using the software \verb|FSL|  \cite{FSL}. Subsequently, the maps from all the subjects were merged into a single four-dimensional image.\\

The statistical validation of our results employed the \verb|randomise| technique \cite{winkler2014permutation} as implemented in \verb|FSL|. This permutation-based strategy requires only minimal assumptions for validity (in our use case, exchangeability of the subjects and a symmetric distribution of the centred maps at a voxel level) and accounts for multiple comparisons when testing voxel by voxel \cite{nichols2002nonparametric}. Using this approach, we conducted a one sample t-test with variance smoothing ($\sigma = 6$mm) and 10000 permutations. Threshold-Free Cluster Enhancement \cite{smith2009threshold} was used, and the  the family-wise error rate was controlled. This procedure yields a three-dimensional brain map of p-values, where each voxel indicates the probability that the balanced accuracy measured at said voxel across different subjects is statistically greater than chance. We chose a significance threshold of $\alpha=0.05$.\\

We also used \verb|randomise| to detect above-chance decoding for each subject independently, identifying informative clusters on an individualized basis. Although \cite{nichols2002nonparametric} advises against using this method on time-correlated data from the same subject, we can safely use it here because we are permuting accuracy values obtained in different data partitions with no time-series structure. 

\section{Experiments}

\subsection{Comparison of DA techniques}

\begin{figure}
    \centering
    \includegraphics[width=\textwidth]{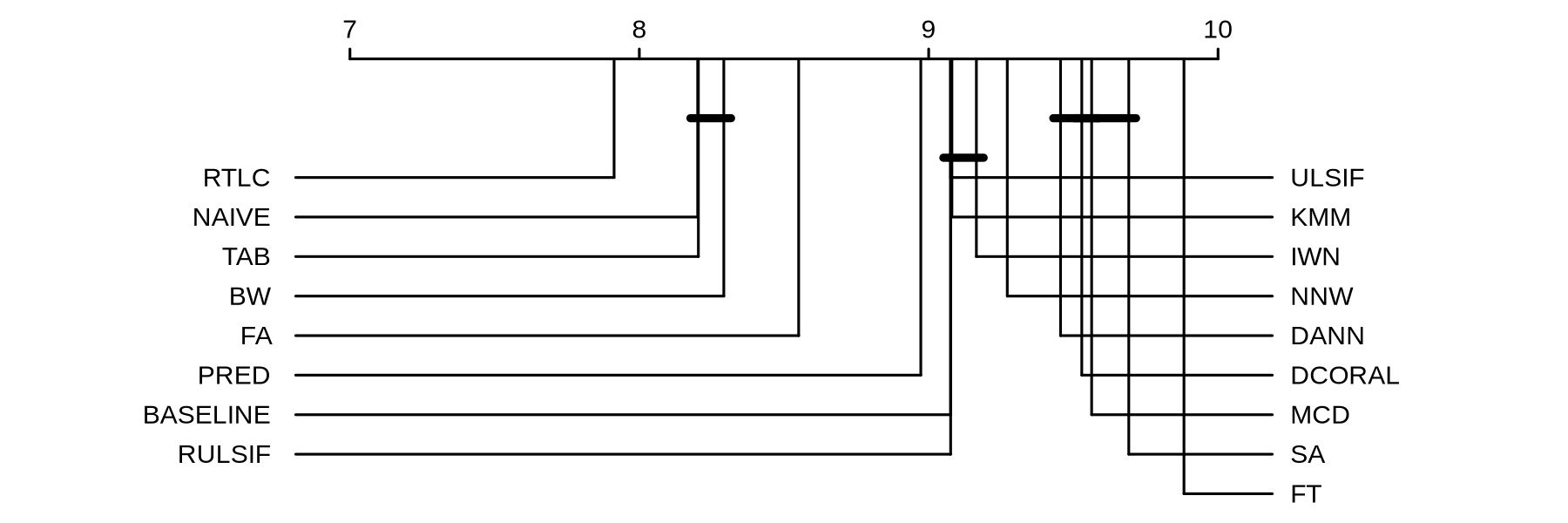}
    \caption{Critical Difference diagram. The scale above shows the average ranking of each algorithm across all observations, and the horizontal lines group together algorithms that are not statistically different.}
    \label{fig:cddiagram}
\end{figure}

Figure \ref{fig:cddiagram} shows the unique Critical Difference (CD) diagram obtained by concatenating all the observations (both across $N_t$ and across different subjects). The scale above shows the average ranking of each algorithm across all observations, and the horizontal lines group together algorithms that are not statistically different. From this, we conclude that RTLC was the best algorithm, followed by the group of NAIVE, TAB, BW and FA which are statistically indistinguishable ($\alpha=0.05$). On the other hand, FT is shown to be the worst.\\

\begin{figure}
    \centering
    \includegraphics[width=\textwidth]{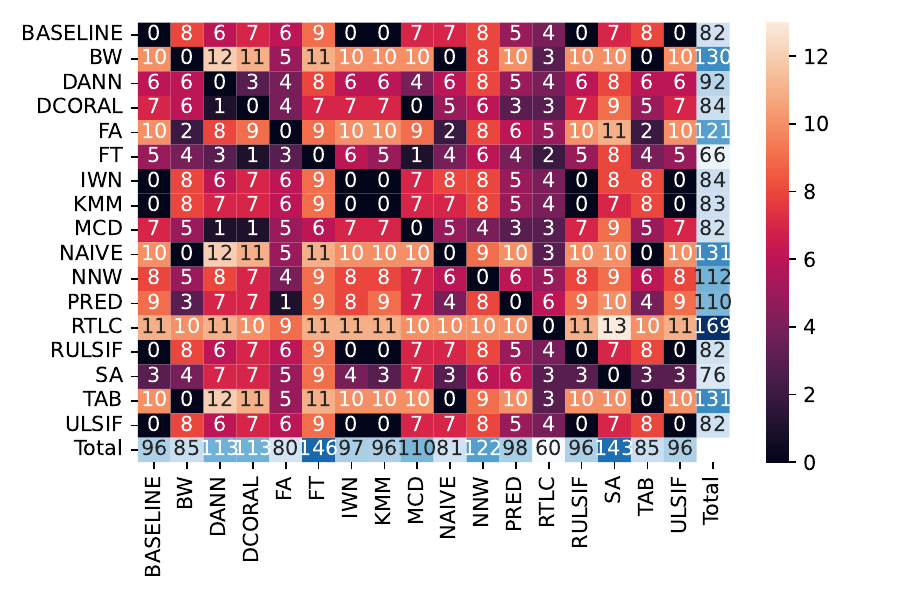}
    \caption{Frequency table. For each cell $(i,j)$, the numerical value shows for how many subjects algorithm $i$ was significantly superior to algorithm $j$. The sum of each row accounts for the number of times each algorithm was significantly better than others, and the sum of each column shows how many times each algorithm was significantly worse than others. }
    \label{fig:freqcd}
\end{figure}

Figure \ref{fig:freqcd} summarizes the results of the tests in the 18 CD diagrams arising from concatenating the 10 tables corresponding to different $N_t$ values for each subject, which are shown in Supplementary Figure \ref{fig:suppCD}. For each cell $(i,j)$ in this table, we report the number of subjects for which algorithm $i$ was significantly better than algorithm $j$ ($\alpha=0.05$). This can be extracted from the CD diagrams by counting how many times $i$ appears to the left of $j$ and subtracting the number of times they are joined by a horizontal line. The sum of each row represents the number of times each algorithm was better than others and, complementarily, the sum of each column reports the number of times each algorithm was significantly worse than others. \\

From Figure \ref{fig:freqcd}, we extract the following highlights:\\

\begin{itemize}
    \item RTLC outperformed others a total of 169 times, and was outperformed 60 times. It was the most outperforming and least outperformed algorithm.
    \item RTLC was better than BASELINE for 11 subjects, worse than BASELINE for 4 and statistically the same for the remaining 3.
    \item RTLC was better than NAIVE for 10 subjects, worse than NAIVE for 4 and statistically the same for the remaining 4.
    \item NAIVE was better than BASELINE for 10 subjects and worse for 7.
    \item FT was the most outperformed and least outperforming algorithm.
\end{itemize}

\FloatBarrier 

\subsection{Analysis of individual ROIs using RTLC} \label{indivrois}

\begin{figure}
    \centering
    \includegraphics[width=0.7\linewidth]{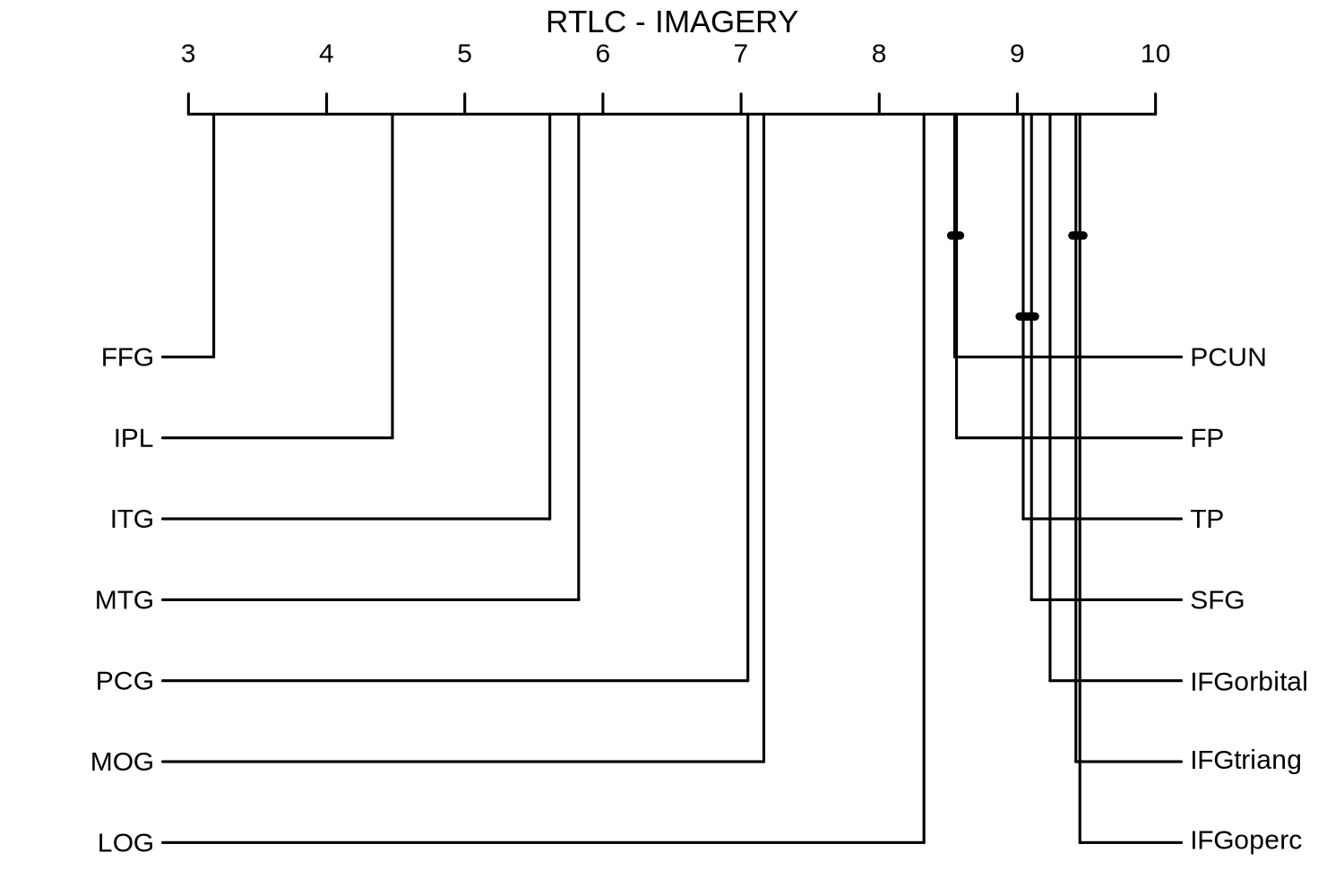}
    \caption{CD diagram of RTLC performance in the imagery domain across individual ROIs. FFG: Fusiform Gyrus; LOG: Lateral occipital gyrus; ITG: Inferior temporal gyrus; TP: Temporal pole; PCG: Posterior cyngulate gyrus; PCUN: Precuneus; IPL: Inferior parietal lobe; MTG: Middle temporal gyrus; SFG: Superior frontal gyrus; IFGoperc: Inferior frontal gyrus, pars opercularis, IFGtriang: Inferior frontal gyrus, pars triangularis; IFGorbital: Inferior frontal gyrus, pars orbitalis; FP: Frontopolar cortex; MOG: Medial orbital gyrus.}
    \label{fig:CDROIS_RTLC}
\end{figure}

Figure \ref{fig:CDROIS_RTLC} shows the CD diagram of the RTLC performance in the imagery domain across individual ROIs, revealing the most informative regions for this method to be the fusiform gyrus (FFG), the inferior parietal lobe (IPL), the inferior temporal gyrus (ITG) and the middle temporal gyrus (MTG). \\

\begin{figure}
    \centering
    \includegraphics[width=0.7\linewidth]{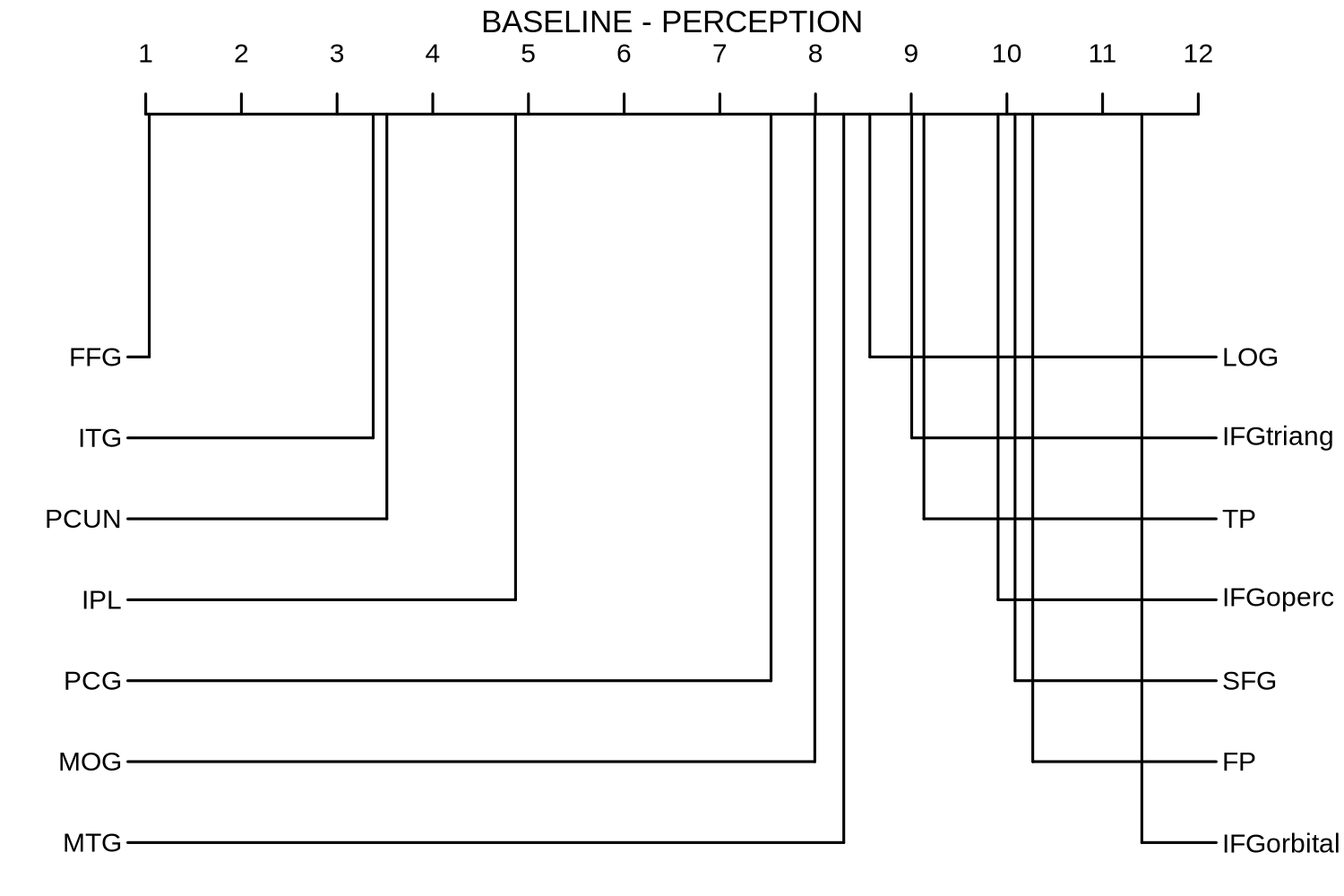}
    \caption{CD diagram of baseline performance in the perception domain across individual ROIs. FFG: Fusiform Gyrus; LOG: Lateral occipital gyrus; ITG: Inferior temporal gyrus; TP: Temporal pole; PCG: Posterior cyngulate gyrus; PCUN: Precuneus; IPL: Inferior parietal lobe; MTG: Middle temporal gyrus; SFG: Superior frontal gyrus; IFGoperc: Inferior frontal gyrus, pars opercularis, IFGtriang: Inferior frontal gyrus, pars triangularis; IFGorbital: Inferior frontal gyrus, pars orbitalis; FP: Frontopolar cortex; MOG: Medial orbital gyrus.}
    \label{fig:CDROIS_BASEPERC}
\end{figure}

Similarly, Figure \ref{fig:CDROIS_BASEPERC} highlights the fusiform gyrus as the most informative region in perception according to the baseline, with an average ranking of 1, meaning that it was the most informative region for all subjects and data partitions.\\

\begin{figure}
    \centering
    \includegraphics[width=0.7\linewidth]{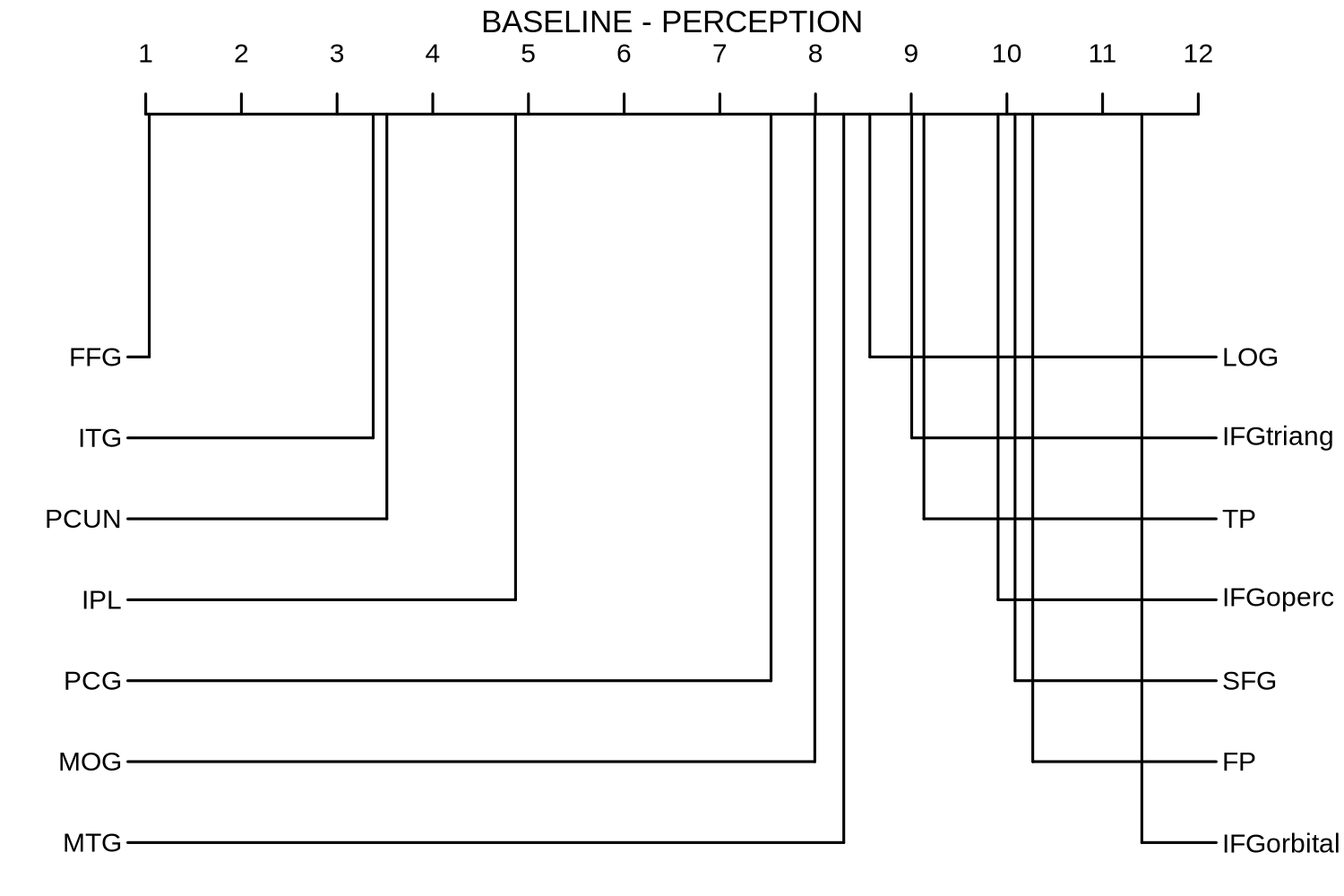}
    \caption{CD diagram of baseline performance in the imagery domain across individual ROIs. FFG: Fusiform Gyrus; LOG: Lateral occipital gyrus; ITG: Inferior temporal gyrus; TP: Temporal pole; PCG: Posterior cyngulate gyrus; PCUN: Precuneus; IPL: Inferior parietal lobe; MTG: Middle temporal gyrus; SFG: Superior frontal gyrus; IFGoperc: Inferior frontal gyrus, pars opercularis, IFGtriang: Inferior frontal gyrus, pars triangularis; IFGorbital: Inferior frontal gyrus, pars orbitalis; FP: Frontopolar cortex; MOG: Medial orbital gyrus.}
    \label{fig:CDROIS_BASEIMAG}
\end{figure}

For completeness, Figure \ref{fig:CDROIS_BASEIMAG} shows the ranking of ROIs acording to the baseline in the imagery domain. Although the fusiform gyrus remains the most informative ROI in this domain, its average ranking has increased, indicating subject variability in the imagery decoding across individual ROIs.\\

In the case of the ``Generic Object Decoding'' dataset, the best and worst methods were respectively PRED and DCORAL (Supplementary Figure \ref{fig:cddiagramGOD}). FA, BW and TAB were also better than NAIVE, although with no statistical significance. PRED outperformed other algorithms 47 times and was only outperformed once (Supplementary Figure \ref{fig:freqcdGOD}). As detailed in \ref{sec:appendixGOD}, the statistical power of the tests was diminished because of the lower number of subjects and feasible data partitions. Although no statistical significance was obtained, Supplementary Figure \ref{fig:suppCDGOD} reveals that FA and PRED were better than NAIVE in 4 subjects, BW for 3 and TAB for 2.

\FloatBarrier 
\subsection{DA-enhanced searchlight analysis}

\begin{figure}
    \centering
    \includegraphics[width=\textwidth]{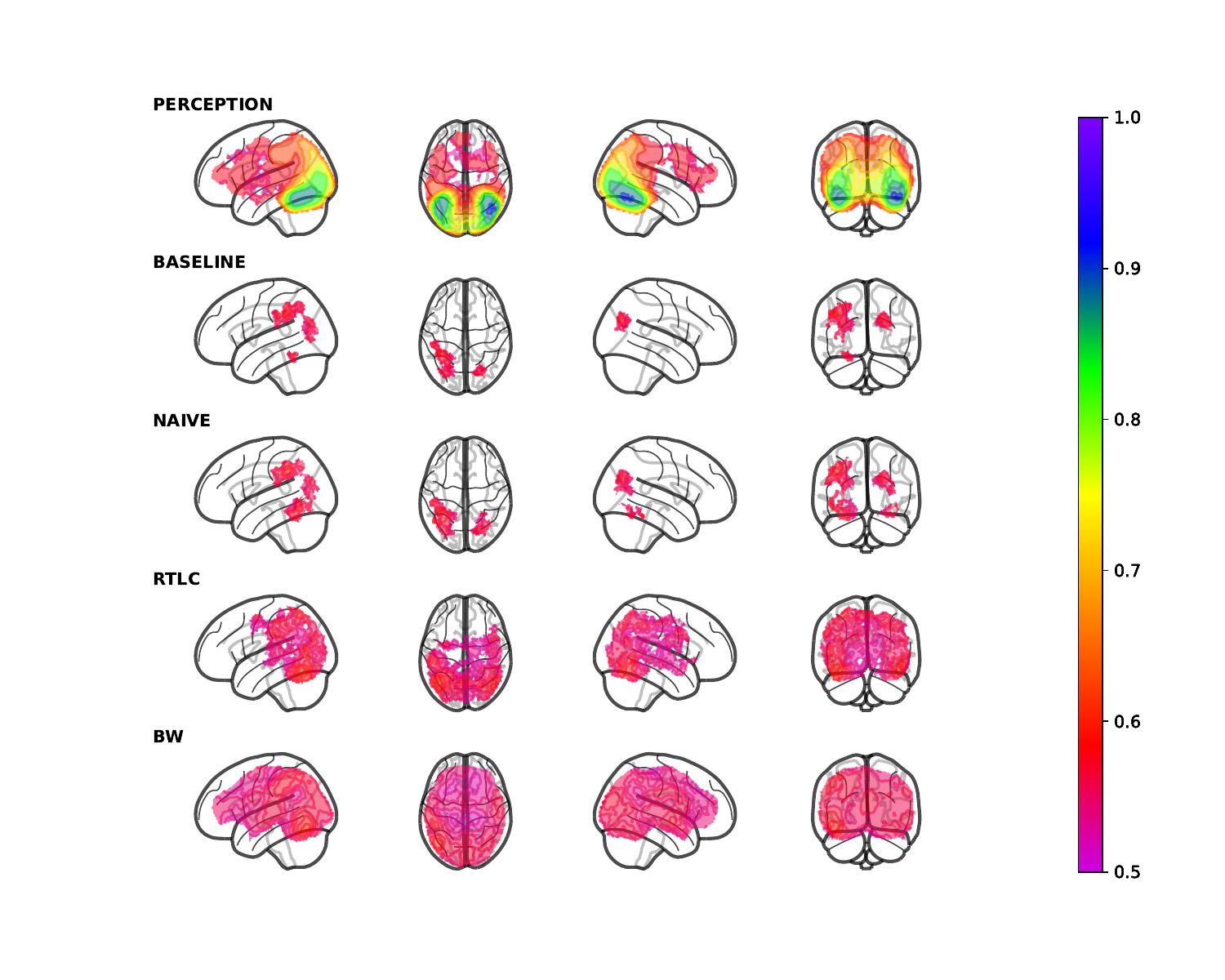}
    \caption{Between-subject average of the balanced accuracy of the different approaches in the regions that obtained statistically above chance decoding with RTLC.}
    \label{fig:projected_acc}
\end{figure}

Figure \ref{fig:projected_acc} shows the projection of the average balanced accuracy of each approach (computed across the subjects) onto the areas of statistically significant brain decoding. We see that only within-domain prediction attains good balanced accuracy when averaged across all the subjects, and that in this case (perception) decoding is possible in a highly distributed set of brain regions. For the baseline and naïve approaches, decoding is only possible in a small, isolated region. On the contrary, our DA-enhanced searchlight with either RLTC and BW is able to obtain significantly above chance imagery decoding in an extensively spread brain volume.\\

\begin{figure}
    \centering
    \includegraphics[width=\textwidth]{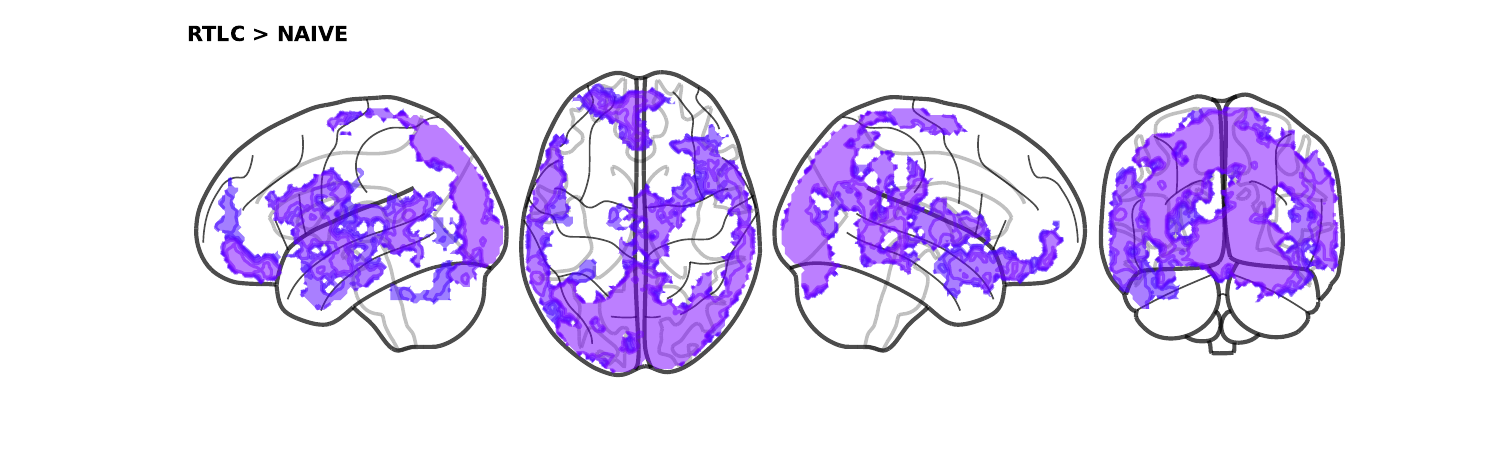}
    \caption{Brain regions where RTLC decoding is significantly better than the naïve approach ($p<0.05$, whole-brain corrected).}
    \label{fig:projected_sign}
\end{figure}

Figure \ref{fig:projected_sign} shows the results of the statistical comparison between RTLC and the naïve approach using a two-sample permutation test, highlighting only the statistically significant clusters ($p<0.05$, whole-brain corrected). The naïve method was never statistically significantly above the baseline. Supplementary Figure \ref{fig:projected_sign_BW} compares BW and naïve. Remarkably, a two-sample permutation test reveals that all the significant clusters that were found with RTLC are still present with BW, highlighting the robustness of the neuroscientific conclusions. \\

\begin{figure}
    \centering   \includegraphics[width=\textwidth]{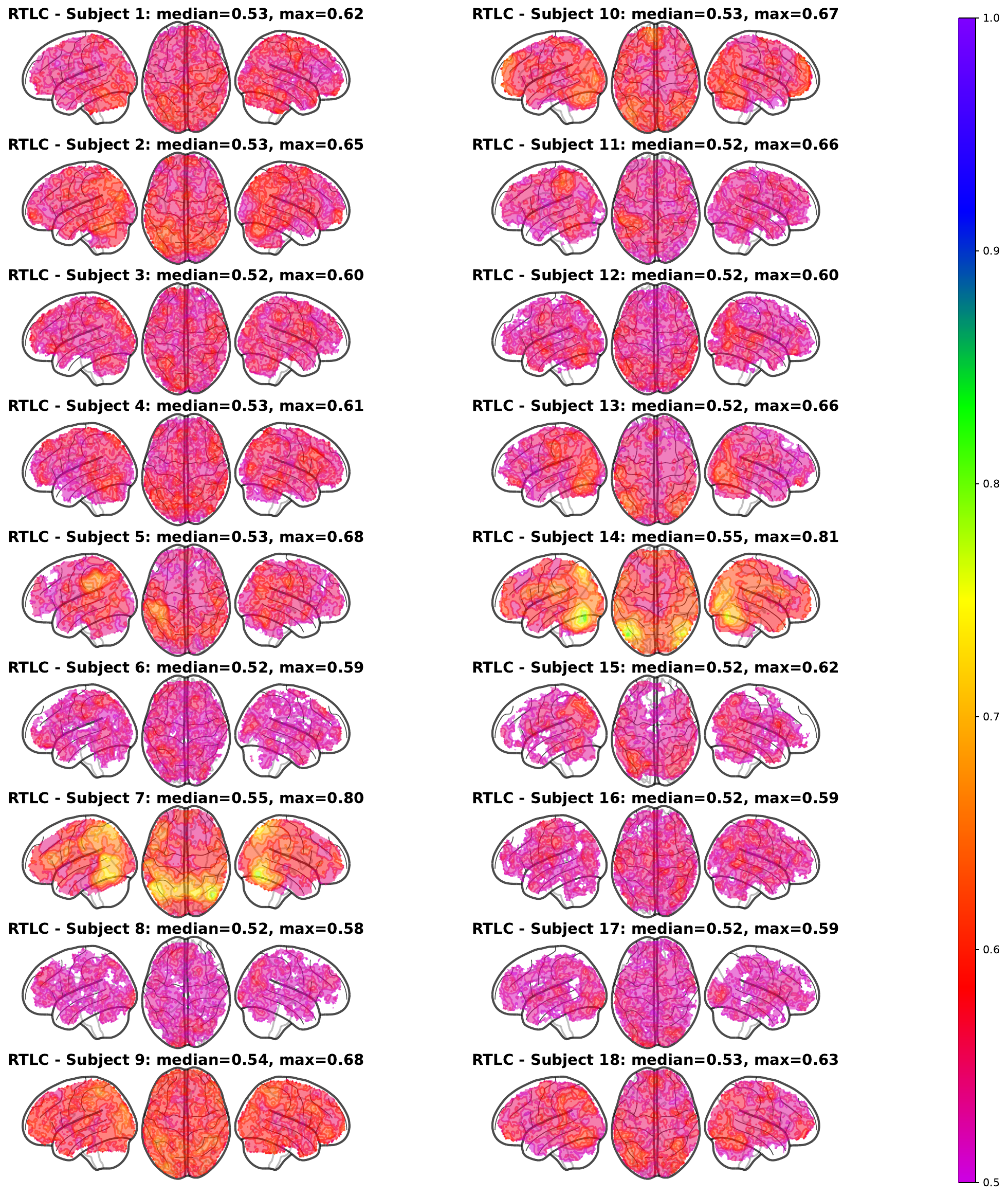}
    \caption{Inter-subject variability of the DA-enhanced searchlight accuracy. For each subject, the colour of each voxel represents the average RTLC accuracy over the 100 repetitions of the experiment in the regions where such accuracy is statistically above 0.5.}
    \label{fig:subj_var}
\end{figure}

Figure \ref{fig:subj_var} accounts for the inter-subject variability of RTLC performance. Conversely, Supplementary Figure \ref{fig:subj_var_BW} displays BW performance for each subject. Each set of brain slices corresponds to a different subject, and shows the average accuracy across the 100 data partitions projected onto the space where such average is statistically greater than chance level. Supplementary Figures \ref{fig:subj_var_perc}, \ref{fig:subj_var_base} and \ref{fig:subj_var_naive}, respectively, show the subject variability of the perception decoding, the baseline cross-domain decoding and the cross-domain decoding with naïve inclusion of target instances in the training set.\\

As illustrated in Supplementary Figure \ref{fig:projected_acc_BW}, the cross-subject average accuracy achieved using BW significantly exceeded chance levels across an even broader range of brain regions compared to RTLC. Notably, with BW, our DA-enhanced searchlight successfully classifies imagery of Living versus Non-Living items in regions that extend into the frontal lobe. However, in terms of symmetry between the left and right hemispheres, the accuracy map for BW exhibits less balance compared to RTLC. This asymmetry is likely attributable to cross-subject averaging, as indicated by the comparison between Figure \ref{fig:subj_var} and Supplementary Figure \ref{fig:subj_var_BW}. Specifically, BW demonstrates poor decoding performance for subjects 9, 15, and 17, whereas RTLC achieves successful decoding for these individuals.

\FloatBarrier 

\section{Discussion}

Although many neuroscience studies face challenges due to distribution shifts, most current analyses of cross-domain experiments in cognitive neuroscience do not address this issue. This oversight hinders the discovery of shared neural representations and limits the applicability of classifiers trained on source domain data to target domains. For instance, in decoded neurofeedback studies, the aim is to induce a specific activation pattern in the brain, usually without participant awareness of the goal state \cite{watanabe2017advances}.  This procedure involves decoding the participant's cognitive state in real time using a previously trained classifier.  Then, a measure of  the similarity between the current brain state and the goal activation pattern is computed, and the degree of similarity is presented as feedback to the subject. Several decoded neurofeedback works involve imagery decoding based on a stimulus presentation \cite{margolles2024unconscious,andersson2019visual} or motor task \cite{de2022one,sepulveda2016feedback}, but none of them implement DA techniques to boost feedback accuracy.\\

We hypothesized that predictive models trained with fMRI data from different experimental conditions could benefit from DA. Our results support this hypothesis, showing that DA significantly improves cross-domain brain decoding from perception to imagery compared to the naive inclusion of imagery instances in the training set. On an extensive ROI level, statistically significant improvement was attained for RTLC in the binary dataset and for PRED in the multiclass setting. Furthermore, 6 out of the 16 DA methods studied significantly outperformed the baseline in our dataset (see Figure \ref{fig:cddiagram}), and 8 methods significantly outperformed the baseline in the multiclass dataset (Supplementary Figure \ref{fig:cddiagramGOD}). \\

For both datasets, BW and TAB were statistically equivalent to NAIVE in the ROI-based comparison, suggesting that these methods may perform even better at a local level, where the curse of dimensionality is absent. Indeed, the DA-enhanced searchlight analysis using BW reveals the potential of this approach compared to NAIVE. Conversely, TAB is less suitable for DA-enhanced searchlight analysis due to its iterative nature and high computational cost. Given the unique characteristics of each dataset, we emphasize the importance of comparing multiple DA methods to guide the selection of the most effective technique for DA-enhanced searchlight, while considering both computational resources and the possibility that DA methods may perform optimally at a local level rather than at the ROI level.\\

With our DA-enhanced searchlight approach, we demonstrated statistically significant pattern similarities between perception and imagery in the visual cortex and beyond, including a distributed set of brain regions in the frontoparietal cortex.\\

Assessing the effect of DA on predictive ability offers a more complete perspective on the role of imagery versus perception in the brain representation of specific concepts. If DA is able to improve classification, it indicates that the initial differences in the distributions corresponding to each stimulus are sufficiently marked to prevent the baseline model from exploiting all the information contained in the data. Furthermore, when the model has a goal beyond testing hypotheses about neural activity, predictive improvement may be critical to the applicability of the classifier, for instance in neurofeedback protocols for clinical interventions. \\

Our results show that DA can be beneficial for cross-domain brain decoding when employed in the union of the ROIs under study, and also when integrated into a searchlight procedure. These results align with previous work stating that BCIs can be improved by incorporating DA into their training methodology \cite{he2020different,phunruangsakao2022deep}. This requires little extra effort, aside from gathering a small sample of data from the target domain. In the case of fMRI studies, acquiring this data involves no significant additional cost. Importantly, we have shown that naïvely concatenating this data into the training set of a conventional classifier is less effective than using DA methods to incorporate the same information.\\

As mentioned above, the best performing DA method for our dataset was RTLC. In the analysis across individual ROIs presented in Section \ref{indivrois}, we assess the contribution of the different ROIs to RTLC performance, as well as their informativeness according to the baseline. We see that RTLC performs best in the FFG, but is also able to extract information from ROIs that are indirectly related to vision such as the inferior parietal lobe (IPL), the inferior temporal gyrus (ITG), and the middle temporal gyrus (MTG).\\

The success of DA is strongly limited by the signal-to-noise ratio of the data. When the task of interest is difficult in the source domain itself, resulting in unreliable starting models, generalization to the target domain will be poor regardless of the DA technique employed. In our case, not all ROIs contain valuable information on visual object category, which becomes evident in Figure \ref{fig:CDROIS_BASEPERC} and in the searchlight analysis. Specifically, Figure \ref{fig:CDROIS_BASEPERC} highlights the fusiform gyrus (FFG) as the most informative ROI for visual perception decoding in all cases, whereas ROIs such as the inferior frontal gyrus, pars orbitalis (IFGorbital), the frontopolar cortex (FP) and the superior frontal gyrus (SFG), among others, encode little information about visual object category. Equivalently, from Supplementary Figure \ref{fig:subj_var_perc} we conclude that all subjects have good perception-to-perception decoding in the areas of the occipital lobe associated with vision, but only some subjects encode visual information in additional brain regions. For around half of the subjects, the visual information is distributed on the whole brain above chance level, even though the decoding is always stronger in the occipital lobe. Hence, the between-subject average decoding is predominant in the occipital lobe but also extends to other areas (Figure \ref{fig:projected_acc} - perception).  \\

When it comes to the direct applicability of standard perception-based models to the imagery domain, Figure \ref{fig:projected_acc} (baseline and naïve) reveals a poor transferability on average. Inspecting Supplementary Figures \ref{fig:subj_var_base} and \ref{fig:subj_var_naive}, we discover that the extent of the domain shift (indicated by the performance gap of these approaches compared to perception prediction) is in fact very subject-dependent. Participants such as Subject 7 and Subject 14 exhibit good transferability, but others such as Subject 5 have sparse above-chance decoding regions in imagery. Crucially, we reiterate that the across-subject average of the decoding is limited to a very small brain region, which makes the standard models unreliable for additional out-of-sample subjects in the imagery domain. \\

In contrast, even when averaged across the subjects, DA-enhanced models attain above chance decoding in mostly the same areas as perception prediction does, although, expectedly, the accuracy drops considerably. Once again, this is explained by the subject variability depicted in Figure \ref{fig:subj_var} and Supplementary Figure \ref{fig:subj_var_BW}. Some subjects benefit from DA more than others do. Participants such as Subject 17 remain a challenge, whereas for others like Subject 5 the adapted models are well tailored to the imagery scenario, highlighting imagery-specific brain regions. In cases such as Subject 5, Subject 13 and Subject 15 the parietal lobe appears to play no role on encoding visual information that distinguishes between the living and non-living classes (see Figure \ref{fig:subj_var_perc}), but it contains relevant imagery information which is discriminative of those classes.\\

Regarding why RTLC and PRED outperform other DA methods in the binary and multiclass datasets respectively in the ROI-based comparison, the simplicity of these techniques might make them adequate to deal with high-dimensional data.
In both datasets, our comparison of DA techniques suffers from the curse of dimensionality. The amount of examples is low compared to the amount of voxels (see Table \ref{tab:appdescriptive} for the exact numbers in our dataset). This may be detrimental to more sophisticated DA techniques, but still feasible for RTLC and PRED. In addition, both RTLC and PRED are tailored to the specific estimator $h$ used for the source domain. Explicitly, RTLC works as a regularized maximum likelihood estimator in the target domain penalized by the distance of the parameters to the optimal source domain parameters, and PRED works by augmenting the features of the source estimator $h$ with its own target domain predictions and then fitting the augmented estimator to the target domain. Thus, if $h$ has good performance, it will be easier to obtain a good $h_{RTLC}$ or $h_{PRED}$. For our dataset, this is evidenced in Supplementary Figure \ref{fig:subj_var_perc} and in Figure \ref{fig:subj_var}, which respectively indicate the performance of the searchlight $h$ and $h_{RTLC}$.\\

The BW technique also performed remarkably in the DA-enhanced searchlight, despite not attaining a statistically significant improvement over NAIVE at the ROI level. Once again, there are some similarities between the methodology of RTLC and BW. BW works by fitting $h_DA$ directly on source and target labelled data according to the modified loss $(1-\gamma)\mathcal{L}(h(X_s),y_s)+\gamma \mathcal{L}(h(X_t),y_t)$, where the parameter $\gamma$ controls the trade-off between fitting the source or the target data. In a way, this is related to RTLC because it minimizes a combined source-target loss where the target component can also be considered a penalty term, although RTLC does it in two consecutive steps.\\

In fact, from the point of view of ML, it is important to analyse the limitations of current DA techniques in real-world scenarios. Usually, DA methods are tested considering artificial benchmarks such as Office-31 \cite{office31} and MNIST \cite{MNIST} or real problems with a low dimensionality. fMRI data offers the opportunity to evaluate these methods, elucidating whether the complexity of the methods is translated into an enhanced performance. Therefore, one of the goals of this paper is to evaluate the performance of state-of-the-art DA techniques when dealing with high-dimensional, noisy and complex data. \\

From a more practical perspective, establishing an accurate learner that is capable to generalize from perception to imagery is relevant for many neuroscience applications, including but not limited to BCIs, neurofeedback training, and clinical scenarios. For instance, in rehabilitation settings, decoding brain activity patterns associated with mental imagery and visual perception could be used to develop interventions for individuals with visual impairments or neurological disorders affecting visual processing.\\

Among the limitations of our work, we lack a comprehensive characterization of the 18 participants of our dataset, which was not possible to achieve with the available data. Our results reveal a large amount of subject variability both in terms of the performance of baseline classifiers and the success of DA efforts, which is natural given the individuality of brain processes, but which we are not able to adequately explain. For instance, measuring the subjective vividness of visual imagery as done in \cite{milton2021behavioral} would have allowed us to establish the relationship between individual imagery strategies and the degree of improvement brought by DA. Another limitation is that some DA methods that did not excel at the ROI-level comparison could have been even better than RTLC in the searchlight analysis, which does not suffer from the curse of dimensionality. In fact, this has been the case for BW, which performed even better than RTLC in the DA-enhanced searchlight. However, due to the computationally intensive nature of the searchlight procedure, a comparison of all the DA techniques on a voxel level would have been unattainable. This work provides a novel precedent of fMRI DA-enhanced searchlight for cross-domain brain decoding, while acknowledging that RTLC is not necessarily the best performing DA technique in this context. Nevertheless, sophisticated DA techniques and/or underlying estimators such as neural networks are impractical to integrate into a searchlight procedure that requires fitting tens of thousands of such models for each data partition. Another limitation is the lack of suitable alternative datasets to strengthen the neuroscientific implications of our findings. We were able to demonstrate the benefits of DA for visual perception to imagery transferability in the visual cortex ROI in a publicly available multiclass dataset, but this dataset is not appropriate for DA-enhanced searchlight due to various factors. To begin with, such dataset contains only 5 subjects, which is a small sample size for the permutation tests that we used to quantify and statistically validate the effect of DA across subjects. More importantly, transforming the searchlight results of all subjects to a common space (such as the MNI) is paramount to draw any conclusions, and the necessary transformation matrices to achieve this are not available with the ``Generic Object Decoding'' dataset. From the perspective of cognitive neuroscience, this paper is focused on investigating commonalities between visual perception and imagery, and the available datasets on this matter are scarce. However, the methodology of the DA-enhanced searchlight introduced in this work can be applied to any domain shift problem in the realm of neuroscience, opening the doors for future research.\\

For instance, inter-individual differences in brain structure and functional organization mean that a model trained on fMRI data from one person may not generalize well to others without substantial adjustment. These inter-subject particularities may be more prominent in subjective tasks such as imagery or memory.  These cross-subject domain-shift scenarios represent a critical problem in which DA may prove useful to improve the generalizability and robustness of models trained to decode brain activity patterns in standard anatomical spaces. Handling cross-subject differences often involves transforming data into a common anatomical or functional space (in our case, the MNI space), but even with alignment, inter-subject variability remains a major factor limiting generalization in fMRI studies. Although cross-subject DA is beyond the scope of our paper, extending the DA-enhanced searchlight procedure to this setting would be important in future studies.\\

In fact, our results highlight the subject-dependency of imagery processes, where inter-subject heterogeneity is much more acute than in visual stimuli processing (as proven by comparing Supplementary Figure A.11 against Supplementary Figure A.12). By applying our methodology on each subject independently, we have managed to gain knowledge about each participant's characteristics (Supplementary Figure 9 and Supplementary Figure A.10). Additionally, by transforming the results to MNI space and running permutation-based statistical tests, we have obtained general insights on the brain functioning of our study population.

\FloatBarrier

\section{Conclusions}
In this paper, we have introduced an effective way to integrate DA into the searchlight procedure, significantly enhancing classification performance in cross-domain brain decoding.\\

Through a comprehensive analysis involving data from 14 ROIs and 18 subjects for a classification task distinguishing between living and non-living objects, we demonstrated that several DA techniques outperformed our baseline model. The baseline model was trained solely on perception data and tested on imagery data. Similar results were observed on a publicly available multiclass dataset, highlighting a clear domain shift between neural activities associated with visual perception and imagery.\\

The best performing method on our dataset was RTLC. Utilizing RTLC, we assessed both the nature and location of the domain shift within our DA-enhanced searchlight approach. Remarkably, RTLC enabled above-chance decoding of imagined stimuli in a highly distributed set of brain regions, including the visual cortex and the frontoparietal cortex. This contrasts sharply with non-DA searchlight procedures, which did not achieve comparable performance. To strengthen the robustness of our neuroscientific findings, we replicated the results using a DA-enhanced searchlight approach with an alternative DA technique (BW). \\

These findings are particularly promising for applications in BCIs and neurofeedback training, suggesting that DA techniques like RTLC can substantially improve the decoding of imagined stimuli or any other critical brain state that investigators may seek to induce in decoded neurofeedback studies. This advancement paves the way for more effective and reliable cross-domain brain decoding methods, potentially revolutionizing how neural data is utilized in practical applications.\\
\FloatBarrier

\backmatter

\section*{Declarations}

\begin{itemize}
\item \textbf{Availability of data and materials:} The code for this paper is available at the GitHub repository \href{https://github.com/AlexOlza/DA-enhanced-searchlight}{AlexOlza/DA-enhanced-searchlight}. The binary perception/imagery dataset analysed during the current study is available in the Open Science Framework (OSF) repository, \href{https://osf.io/audmx/}{https://osf.io/audmx/}. The ``Generic Object Decoding'' dataset that supports the findings of this study is available in Figshare with the identifier \href{https://figshare.com/articles/dataset/Generic_Object_Decoding/7387130}{https://figshare.com/articles/dataset/Generic\_Object\_Decoding/7387130}.
\item \textbf{Competing interests:} The authors declare that they have no competing interests.
\item \textbf{Funding:} This work has been supported by the Basque Government via the IKUR strategy,  project IT1504-22 and the BERC 2022-2025 program, by the Spanish Ministry of Science and Innovation (PID2022-137442NB-I00), by Elkartek (KK-2023/00090), by the Spanish Ministry of Economy and Competitiveness, through the ’Severo Ochoa’ Programme for Centres/Units of Excellence (CEX2020-001010-S) and also from project grant PID2019-105494GB-I00. 
\item \textbf{Author's contributions:} AO participated in the conceptualization and research of the methodology, implemented the software and wrote the original draft.
DS was involved in the conceptualization, data curation, funding acquisition, supervision and revision of the manuscript. 
RS contributed to the conceptualization, funding acquisition, supervision and revision of the manuscript.
All authors read and approved the final manuscript.
\item \textbf{Acknowledgements:} We are grateful to the anonymous subjects that took part in experimental data acquisition.
\end{itemize}

\section{List of abbreviations}
\begin{itemize}
    \item DA: Domain Adaptation
    \item MRI: Magnetic Resonance Imaging
    \item fMRI: Functional MRI
    \item MVPA: Multivoxel (or multivariate) Pattern Classification Analysis
    \item BCI: Brain-Computer Interface
    \item ROI: Region of Interest
    \item ML: Machine Learning
    \item EEG: Electroencephalography
    \item CNN: Convolutional Neural Network
    \item MNI: Montreal Neurological Institute
    \item LR: Logistic Regression
    \item CD: Critical Difference (diagram)
    \item FFG: Fusiform gyrus
    \item IPL: Inferior Parietal Lobe
    \item ITG: Inferior Temporal Lobe
    \item MTG: Middle Temporal Gyrus
\end{itemize}

\begin{appendices}

\section{Supplementary Results}

\begin{table}[h!]
    \centering
    \begin{tabular}{l|l}
    Abbreviation& Name \\
    \hline
         FFG & Fusiform gyrus\\
         LOG& Lateral occipital gyrus\\
         ITG&  Inferior temporal gyrus\\
         TP& Temporal pole\\
         PCG& Posterior cingulate gyrus\\
         PCUN& Precuneus\\
         IPL& Inferior parietal lobe\\
         MTG& Middle temporal gyrus\\
         SFG& Superior frontal gyrus\\
         IFGoperc& Inferior frontal gyrus, pars opercularis\\
         IFGorbital& Inferior frontal gyrus, pars orbitalis \\
         IFGtriang& Inferior frontal gyrus, pars triangularis\\
         FP& Frontopolar cortex\\
         MOG& Medial orbital gyrus
    \end{tabular}
    \caption{Abbreviations and names of the ROIs under study}
    \label{tab:rois}
\end{table}

\begin{table}[h!]
\begin{tabular}{lrrrrrr}
\hline
\multicolumn{3}{|l}{} & \multicolumn{2}{|c|}{Perception} & \multicolumn{2}{c|}{Imagery} \\ \hline
\multicolumn{1}{|l|}{Subject} & \multicolumn{1}{r|}{Voxels (ROIs)} & \multicolumn{1}{r|}{Voxels} & \multicolumn{1}{r|}{Instances} & \multicolumn{1}{r|}{Prevalence} & \multicolumn{1}{r|}{Instances} & \multicolumn{1}{r|}{Prevalence} \\ \hline
\multicolumn{1}{|l|}{1} & \multicolumn{1}{r|}{10373}& \multicolumn{1}{r|}{35744} & \multicolumn{1}{r|}{368} & \multicolumn{1}{r|}{0.503} & \multicolumn{1}{r|}{234} & \multicolumn{1}{r|}{0.513} \\ \hline
\multicolumn{1}{|l|}{2} & \multicolumn{1}{r|}{10675}& \multicolumn{1}{r|}{35464} & \multicolumn{1}{r|}{362} & \multicolumn{1}{r|}{0.500} & \multicolumn{1}{r|}{233} & \multicolumn{1}{r|}{0.494} \\ \hline
\multicolumn{1}{|l|}{3} & \multicolumn{1}{r|}{10383}& \multicolumn{1}{r|}{37775} & \multicolumn{1}{r|}{362} & \multicolumn{1}{r|}{0.494} & \multicolumn{1}{r|}{232} & \multicolumn{1}{r|}{0.509} \\ \hline
\multicolumn{1}{|l|}{4} & \multicolumn{1}{r|}{10852}& \multicolumn{1}{r|}{41513} & \multicolumn{1}{r|}{362} & \multicolumn{1}{r|}{0.517} & \multicolumn{1}{r|}{233} & \multicolumn{1}{r|}{0.511} \\ \hline
\multicolumn{1}{|l|}{5} & \multicolumn{1}{r|}{10768} & \multicolumn{1}{r|}{ 40194}& \multicolumn{1}{r|}{362} & \multicolumn{1}{r|}{0.506} & \multicolumn{1}{r|}{233} & \multicolumn{1}{r|}{0.494} \\ \hline
\multicolumn{1}{|l|}{6} & \multicolumn{1}{r|}{10577} & \multicolumn{1}{r|}{41162}& \multicolumn{1}{r|}{364} & \multicolumn{1}{r|}{0.495} & \multicolumn{1}{r|}{233} & \multicolumn{1}{r|}{0.498} \\ \hline
\multicolumn{1}{|l|}{7} & \multicolumn{1}{r|}{11078}& \multicolumn{1}{r|}{47814} & \multicolumn{1}{r|}{364} & \multicolumn{1}{r|}{0.500} & \multicolumn{1}{r|}{231} & \multicolumn{1}{r|}{0.511} \\ \hline
\multicolumn{1}{|l|}{8} & \multicolumn{1}{r|}{11178}& \multicolumn{1}{r|}{ 43787} & \multicolumn{1}{r|}{366} & \multicolumn{1}{r|}{0.511} & \multicolumn{1}{r|}{233} & \multicolumn{1}{r|}{0.502} \\ \hline
\multicolumn{1}{|l|}{9} & \multicolumn{1}{r|}{13916}& \multicolumn{1}{r|}{51748} & \multicolumn{1}{r|}{361} & \multicolumn{1}{r|}{0.504} & \multicolumn{1}{r|}{234} & \multicolumn{1}{r|}{0.491} \\ \hline
\multicolumn{1}{|l|}{10} & \multicolumn{1}{r|}{11841}& \multicolumn{1}{r|}{46884} & \multicolumn{1}{r|}{362} & \multicolumn{1}{r|}{0.497} & \multicolumn{1}{r|}{231} & \multicolumn{1}{r|}{0.506} \\ \hline
\multicolumn{1}{|l|}{11} & \multicolumn{1}{r|}{10807}& \multicolumn{1}{r|}{41465} & \multicolumn{1}{r|}{362} & \multicolumn{1}{r|}{0.503} & \multicolumn{1}{r|}{233} & \multicolumn{1}{r|}{0.511} \\ \hline
\multicolumn{1}{|l|}{12} & \multicolumn{1}{r|}{12171}& \multicolumn{1}{r|}{44995} & \multicolumn{1}{r|}{364} & \multicolumn{1}{r|}{0.500} & \multicolumn{1}{r|}{232} & \multicolumn{1}{r|}{0.496} \\ \hline
\multicolumn{1}{|l|}{13} & \multicolumn{1}{r|}{12208}& \multicolumn{1}{r|}{46920} & \multicolumn{1}{r|}{365} & \multicolumn{1}{r|}{0.496} & \multicolumn{1}{r|}{233} & \multicolumn{1}{r|}{0.506} \\ \hline
\multicolumn{1}{|l|}{14} & \multicolumn{1}{r|}{10823}& \multicolumn{1}{r|}{38441} & \multicolumn{1}{r|}{365} & \multicolumn{1}{r|}{0.496} & \multicolumn{1}{r|}{234} & \multicolumn{1}{r|}{0.500} \\ \hline
\multicolumn{1}{|l|}{15} & \multicolumn{1}{r|}{12127}& \multicolumn{1}{r|}{40111} & \multicolumn{1}{r|}{360} & \multicolumn{1}{r|}{0.500} & \multicolumn{1}{r|}{234} & \multicolumn{1}{r|}{0.491} \\ \hline
\multicolumn{1}{|l|}{16} & \multicolumn{1}{r|}{9161}& \multicolumn{1}{r|}{33991} & \multicolumn{1}{r|}{364} & \multicolumn{1}{r|}{0.489} & \multicolumn{1}{r|}{232} & \multicolumn{1}{r|}{0.500} \\ \hline
\multicolumn{1}{|l|}{17} & \multicolumn{1}{r|}{12179} & \multicolumn{1}{r|}{45649}& \multicolumn{1}{r|}{363} & \multicolumn{1}{r|}{0.510} & \multicolumn{1}{r|}{233} & \multicolumn{1}{r|}{0.511} \\ \hline
\multicolumn{1}{|l|}{18} & \multicolumn{1}{r|}{11214} & \multicolumn{1}{r|}{43522}& \multicolumn{1}{r|}{361} & \multicolumn{1}{r|}{0.507} & \multicolumn{1}{r|}{234} & \multicolumn{1}{r|}{0.504} \\ \hline
\end{tabular}
\caption{Dataset characteristics for each subject}
\label{tab:appdescriptive}
\end{table}
\clearpage
\newgeometry{top=0.2in}
\begin{figure}[p]
    \centering
    \includegraphics[width=\textwidth,height=\textheight]{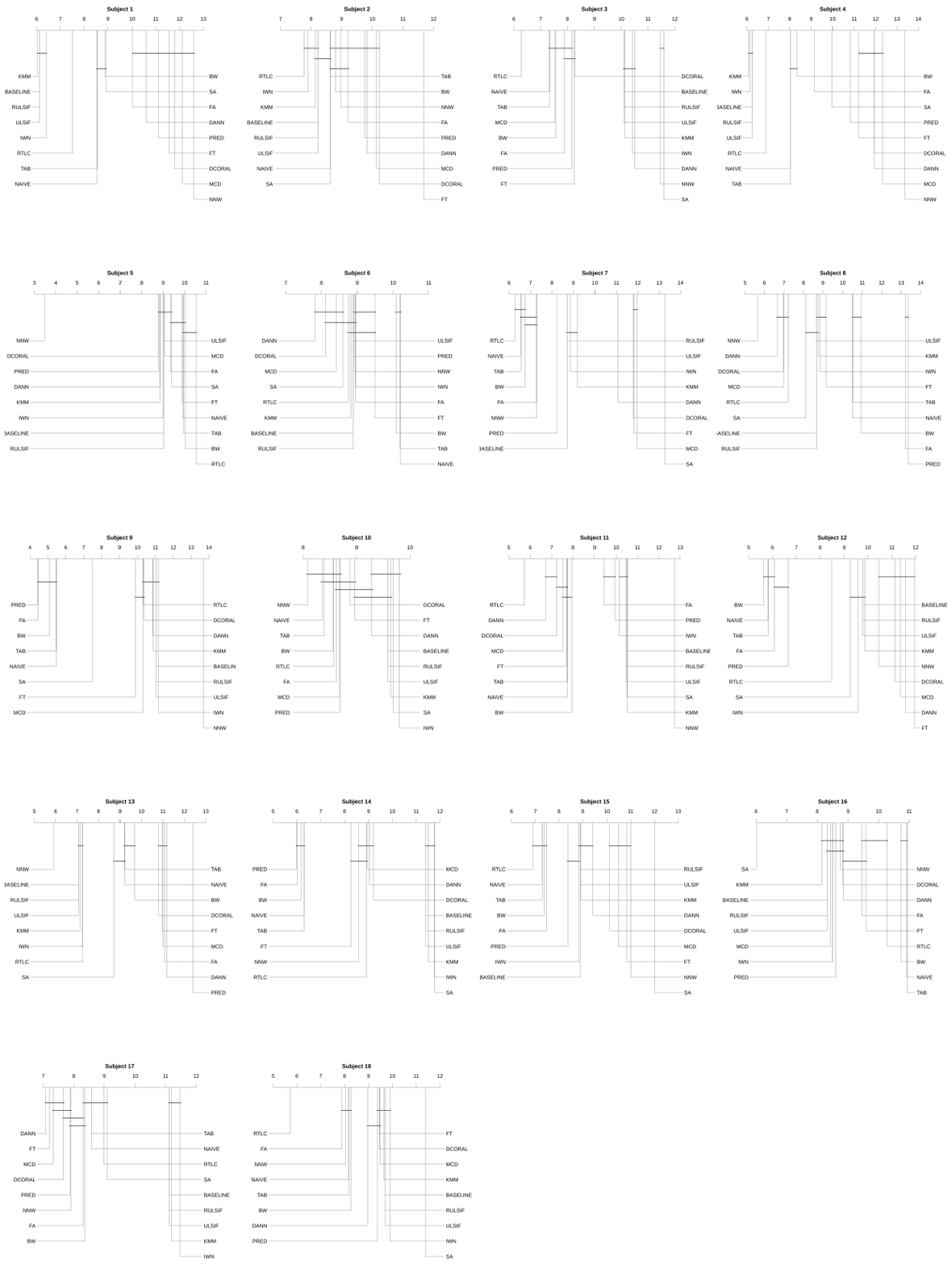}
    \caption{Critical Difference diagrams for each subject, comparing different DA techniques on the union of 14 ROIs.}
    \label{fig:suppCD}
\end{figure}
\restoregeometry
\clearpage
\begin{figure}[p]
    \centering
    \includegraphics[width=\textwidth]{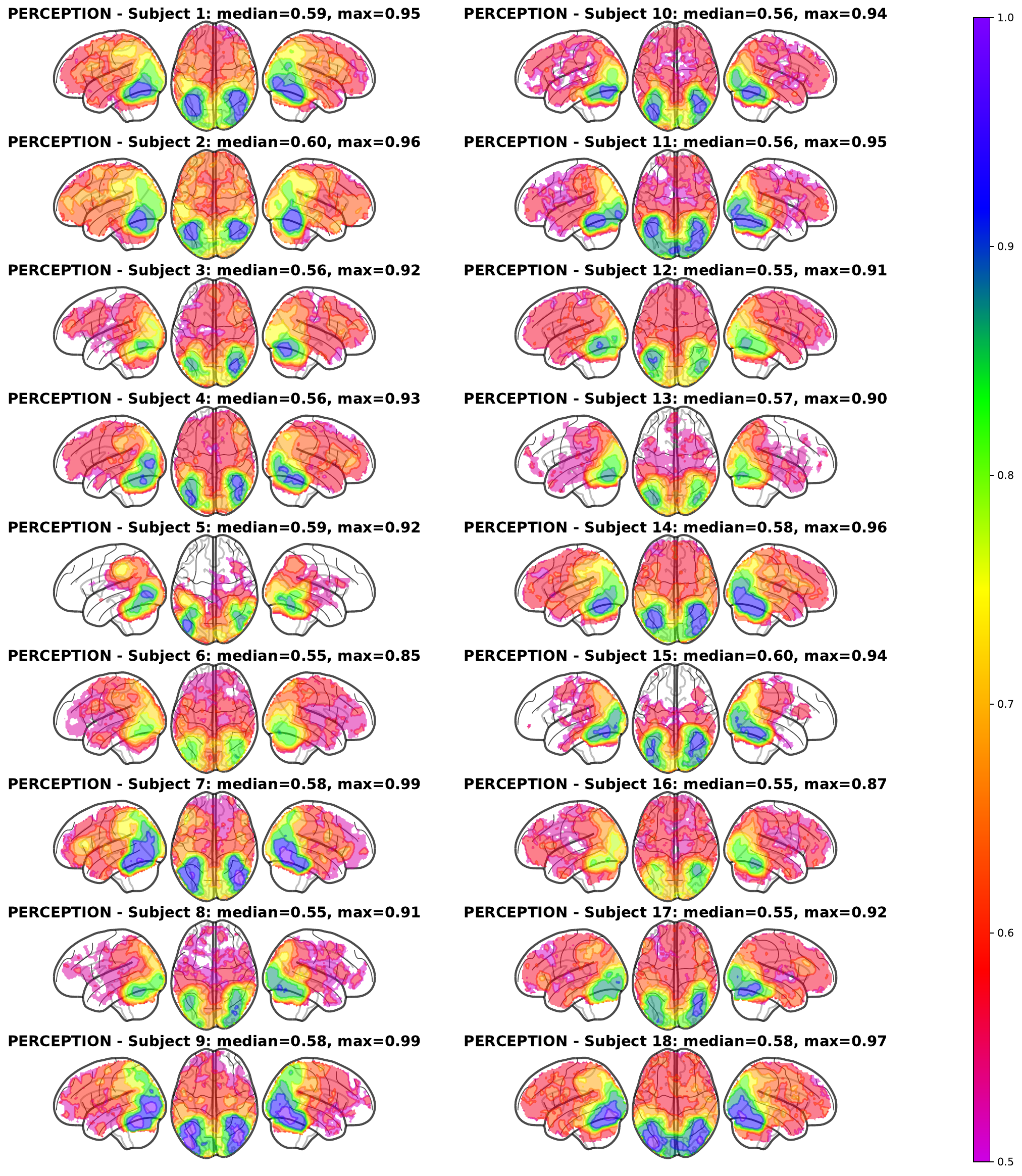}
    \caption{Subject variability of the within-domain searchlight accuracy. For each subject, the colour of each voxel represents the average perception-to-perception accuracy over the 100 repetitions of the experiment in the regions where such accuracy is statistically significantly above 0.5.}
    \label{fig:subj_var_perc}
\end{figure}

\begin{figure}[p]
    \centering
    \includegraphics[width=\textwidth]{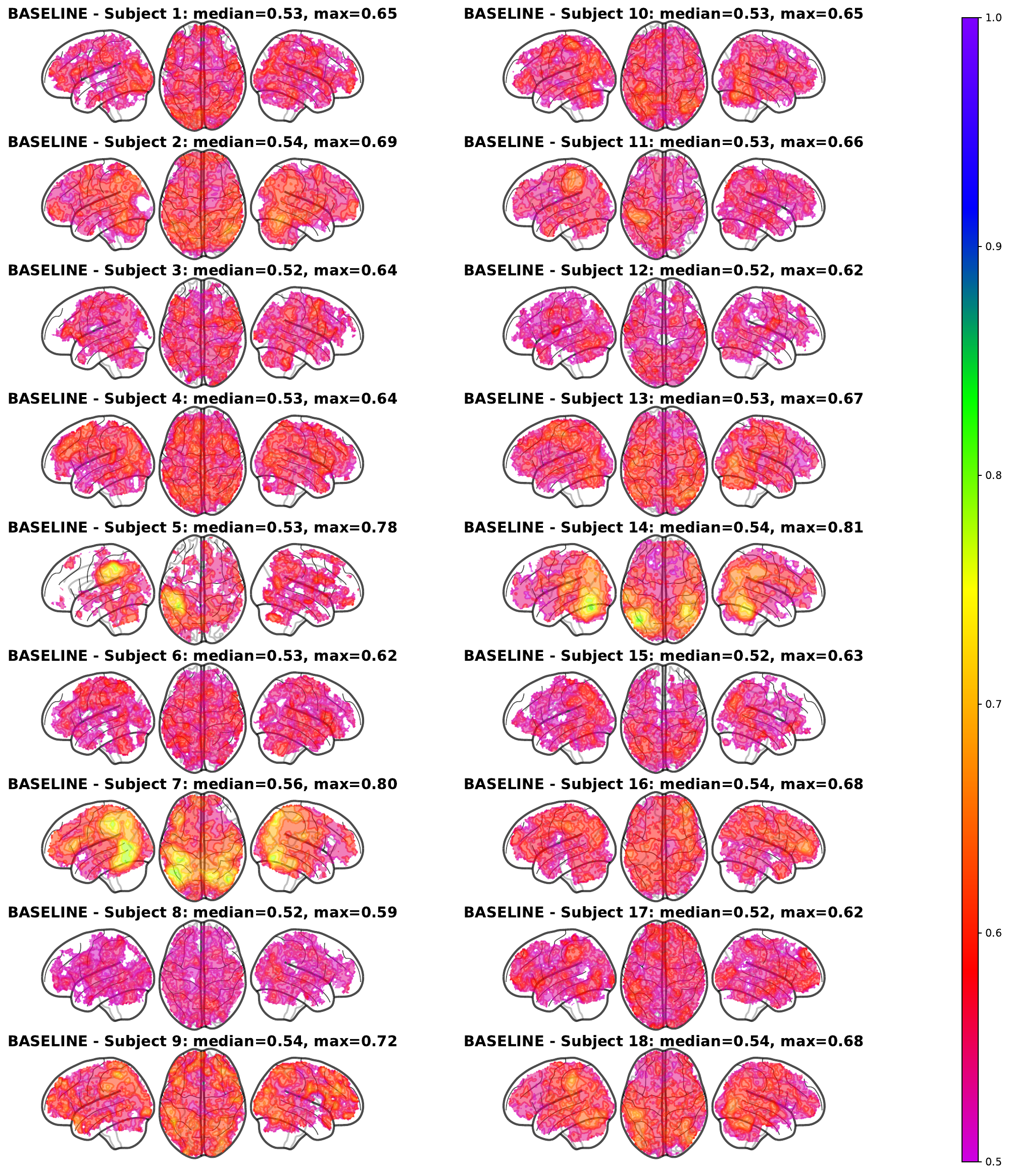}
    \caption{Subject variability of the cross-domain searchlight accuracy. For each subject, the colour of each voxel represents the average baseline perception-to-imagery accuracy over the 100 repetitions of the experiment in the regions where such accuracy is statistically significantly above 0.5.}
    \label{fig:subj_var_base}
\end{figure}

\begin{figure}[p]
    \centering
    \includegraphics[width=\textwidth]{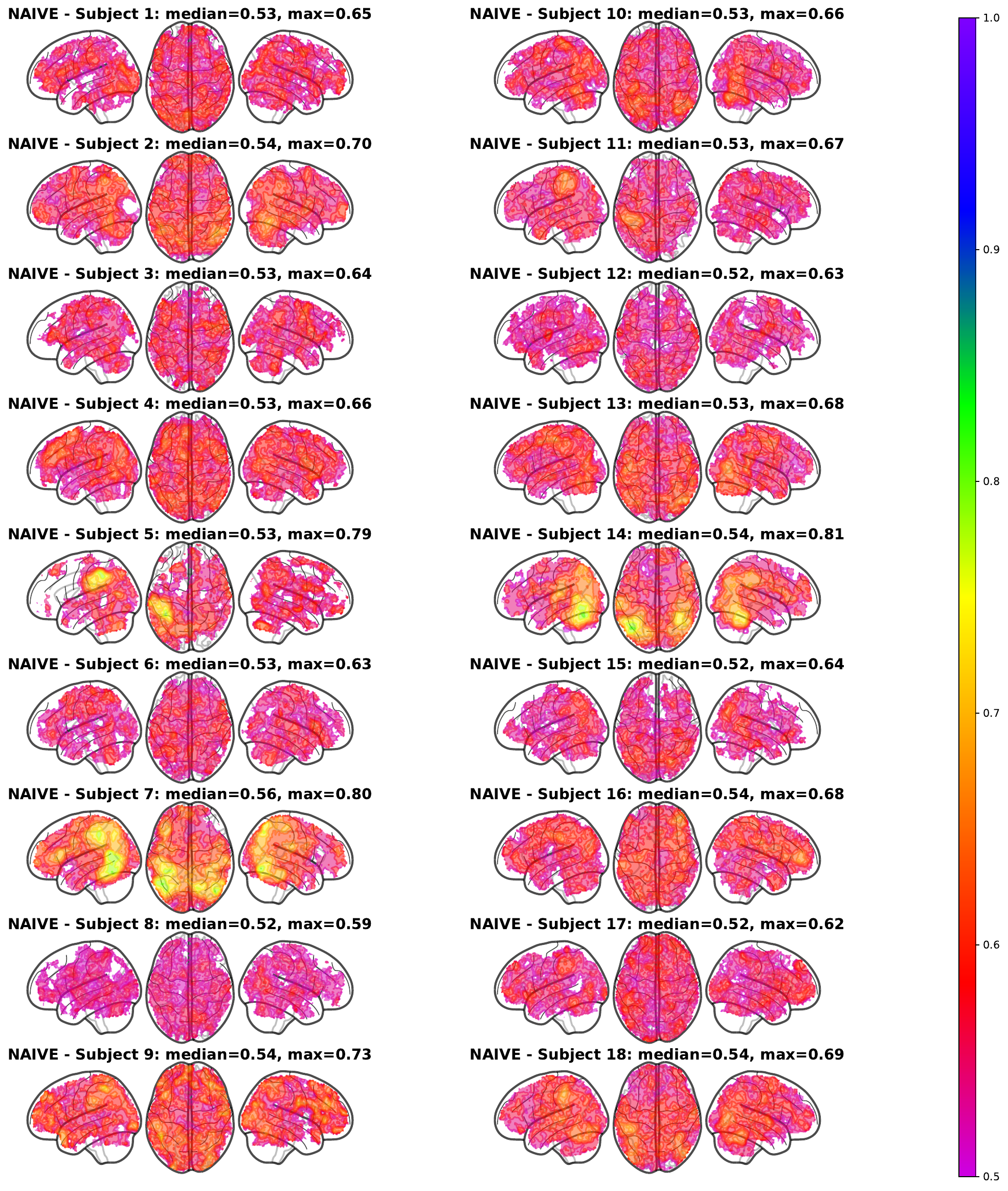}
    \caption{Subject variability of the cross-domain searchlight accuracy with 100 target domain instances available during training. For each subject, the colour of each voxel represents the average accuracy over the 100 repetitions of the experiment in the regions where such accuracy is statistically significantly above 0.5.}
    \label{fig:subj_var_naive}
\end{figure}
\FloatBarrier

\section{Results on ``Generic Object Decoding'' dataset}\label{sec:appendixGOD}

\subsection{Dataset description}
The publicly available ``Generic Object Decoding'' dataset \cite{horikawa2017generic} gathers fMRI scans from 5 healthy subjects that were exposed to visual stimuli from 50 different categories (perception condition) and were later instructed to imagine those categories (imagery condition). Both of these experiments constituted the evaluation data for the methodology proposed in the original paper \cite{horikawa2017generic}. Data acquisition and preprocessing are discussed in \cite{horikawa2017generic}.\\

In the perception condition, each category was presented a total of 35 times and the experiment involved 35 runs. The stimuli presentation order was randomized across runs. Each category included a single representative image, which was extracted from the THINGS \cite{THINGS} database. \\

The imagery condition involved 20 runs, each of which included half of the categories in a semi-random order, such that every two consecutive runs contained all the categories. Each run was composed of 25 imagery events, and therefore, each category was imagined a total of 10 times during the experiment.\\

This dataset allows the opportunity to test the applicability of DA methods to assess the transferability of visual perception decoding to the imagery domain in a multiclass classification setting. However, the reduced amount of imagery data per class (10 examples) is a key limitation for training and evaluation, both of which require all categories to be present in the corresponding data partitions. Furthermore, the need to ensure that examples from the same experimental run are not present in both train and test partitions renders training impossible given the distribution of categories across the runs in the data acquisition step. \\

For the reasons listed above, investigating the impact of DA on this dataset is only feasible if the number of classes is reduced in a way that preserves the correlation between class labels and neural activity. We propose a semantic grouping into 9 broader classes (Table \ref{tab:GODcategories}), one of which must be discarded due to insufficient imagery data (``Weapons", containing only the subclass ``Cannon": 10 examples). This semantic grouping results in an imbalanced dataset, because each broader class contains a different number of subclasses. Many other semantic groupings are possible, as long as they are crafted in such a way that preserves the human interpretation of the stimuli.\\

\begin{table}[h!]
    \centering
    \begin{tabular}{|l|p{4cm}|c|}
        \hline
        \textbf{Broad Category} & \textbf{Original Subclasses} & \textbf{Number of Examples} \\
        \hline
        Animals & Goldfish, Owl, Common iguana, Duck, Swan, Conch, Crab, Killer whale, Leopard, Bat, Housefly, Butterfly, Goat, Camel, Domestic llama & 150 \\
        \hline
        Clothing and Accessories & Cowboy hat, Silk hat, Football helmet, Sock, Welder's mask & 50 \\
        \hline
        Household Items & Beer mug, Bowling ball, Hammock, Mailbox, Umbrella & 50 \\
        \hline
        Musical Instruments & Electric guitar, Grand piano, Harp, Mandolin, Tambourine & 50 \\
        \hline
        Structures and Monuments & Coffin, Gravestone, Minaret, Stained-glass window & 40 \\
        \hline
        Technology and Devices & IPod, Microwave, Videocassette recorder, Washing machine & 40 \\
        \hline
        Tools and Equipment & Fire extinguisher, Knob, Tweezer, Planchet, Shredder & 50 \\
        \hline
        Vehicles and Transportation & Airliner, Barrow, Bulldozer, Canoe, Covered wagon, Snowmobile & 60 \\
        \hline
        Weapons & Cannon & 10 \\
        \hline
    \end{tabular}
    \caption{Broad categories, their original subclasses, and the number of examples.}
    \label{tab:GODcategories}
\end{table}

\FloatBarrier
\subsection{Machine learning task and validation framework}

We performed the comparison of the 15 DA methods under study, as well as the NAIVE and BASELINE methods as defined in Section 4. The validation framework was exactly the same, except for the number of data partitions, which decreased from 100 to 24 as a consequence of requiring all 8 classes to be present for both training and testing. The number of target domain instances in the training set ($N_t$) was also increased to account for the multiclass setting ($N_t= 200, 250, 300$). Random over-sampling was used in both the source and target domains to balance the dataset before training.\\

We used all the voxels from the largest ROI provided by the dataset authors, which covers the Visual Cortex (VC) as defined in \cite{horikawa2017generic} Supplementary Figure 1. Because of the reduced number of subjects and viable data partitions, the total experiments per algorithm were significantly fewer for the ``Generic Object Decoding'' dataset, impacting the power of the statistical tests.\\

\FloatBarrier

\subsection{Comparison of DA techniques}

\begin{figure}
    \centering
    \includegraphics[width=\textwidth]{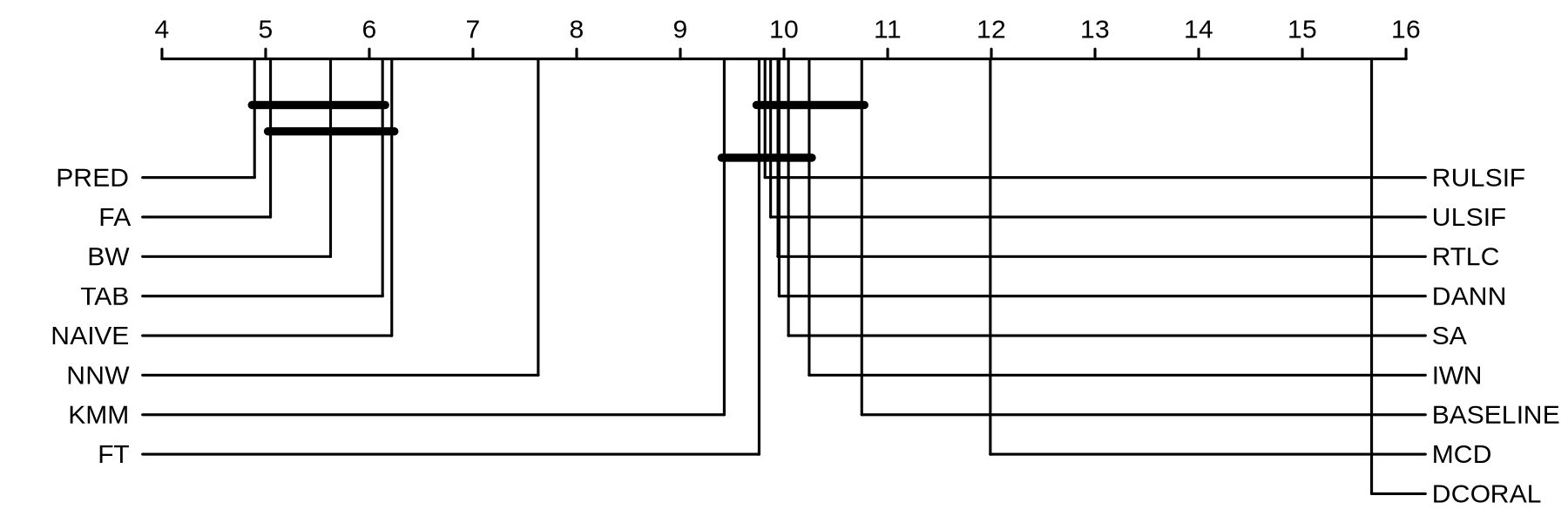}
    \caption{Critical Difference diagram. The scale above shows the average ranking of each algorithm across all observations, and the horizontal lines group together algorithms that are not statistically different.}
    \label{fig:cddiagramGOD}
\end{figure}

From Figure \ref{fig:cddiagramGOD} we conclude that PRED, FA, BW and TAB were the best algorithms. Among those, PRED was statistically significantly better than NAIVE ($\alpha=0.05$). \\

\begin{figure}
    \centering
    \includegraphics[width=\textwidth]{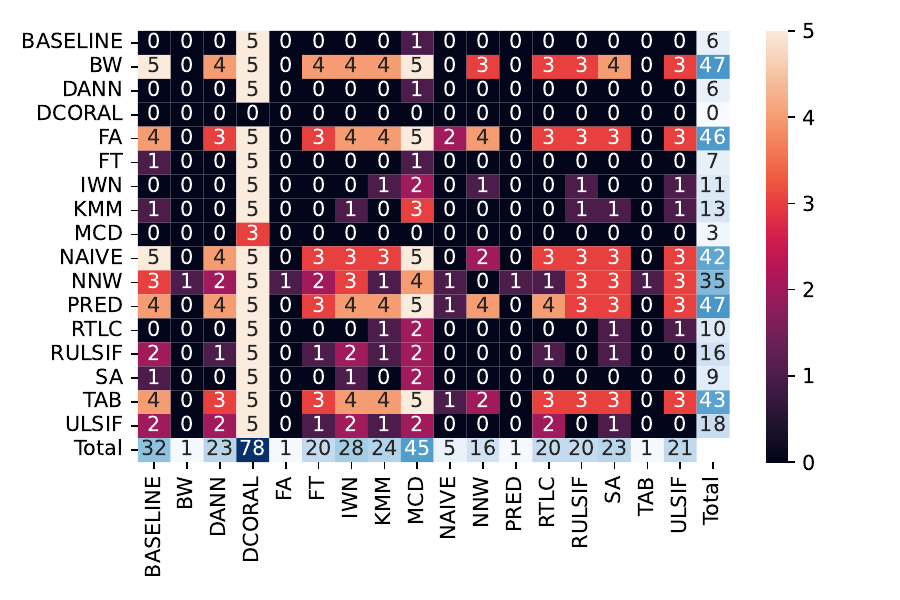}
    \caption{Frequency table. For each cell $(i,j)$, the numerical value shows for how many subjects algorithm $i$ was significantly superior to algorithm $j$. The sum of each row accounts for the number of times each algorithm was significantly better than others, and the sum of each column shows how many times each algorithm was significantly worse than others. }
    \label{fig:freqcdGOD}
\end{figure}

Figure \ref{fig:freqcdGOD} indicates that PRED significantly outperformed other algorithms 47 times, and was only outperformed once (by NNW). It was the most outperforming algorithm. Similarly, FA, BW and TAB were outperformed only once (by NNW). \\

\begin{figure}[p]
    \centering
    \includegraphics[width=\textwidth]{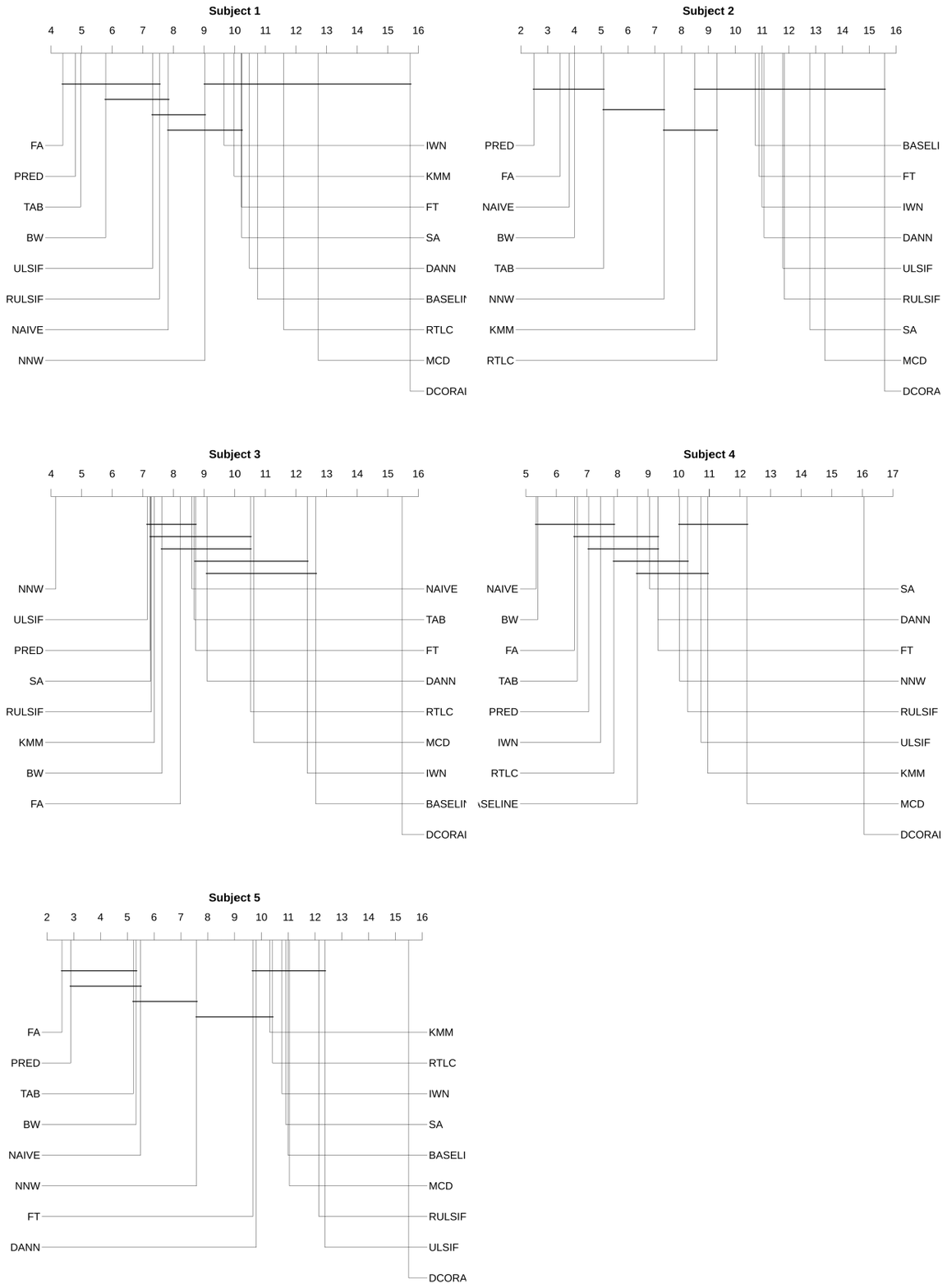}
    \caption{Critical Difference diagrams for each subject, comparing different DA techniques in the Visual Cortex.}
    \label{fig:suppCDGOD}
\end{figure}

Due to the decrease in the number of subjects and feasible data partitions, the total number of experiments per algorithm was dramatically lower for the ``Generic Object Decoding'' dataset, which affects the power of the statistical tests summarized in figures \ref{fig:cddiagramGOD} and \ref{fig:freqcdGOD}. Figure \ref{fig:suppCDGOD} shows the results for each subject, revealing that FA and PRED were better than NAIVE in 4 subjects, BW for 3 and TAB for 2.

\clearpage
\newgeometry{top=0.2in}
\restoregeometry
\clearpage
\FloatBarrier

\section{DA-enhanced searchlight results with the Balanced Weighting technique} \label{sec:appendixBW}

\begin{figure}[h]
    \centering
    \includegraphics[width=\textwidth]{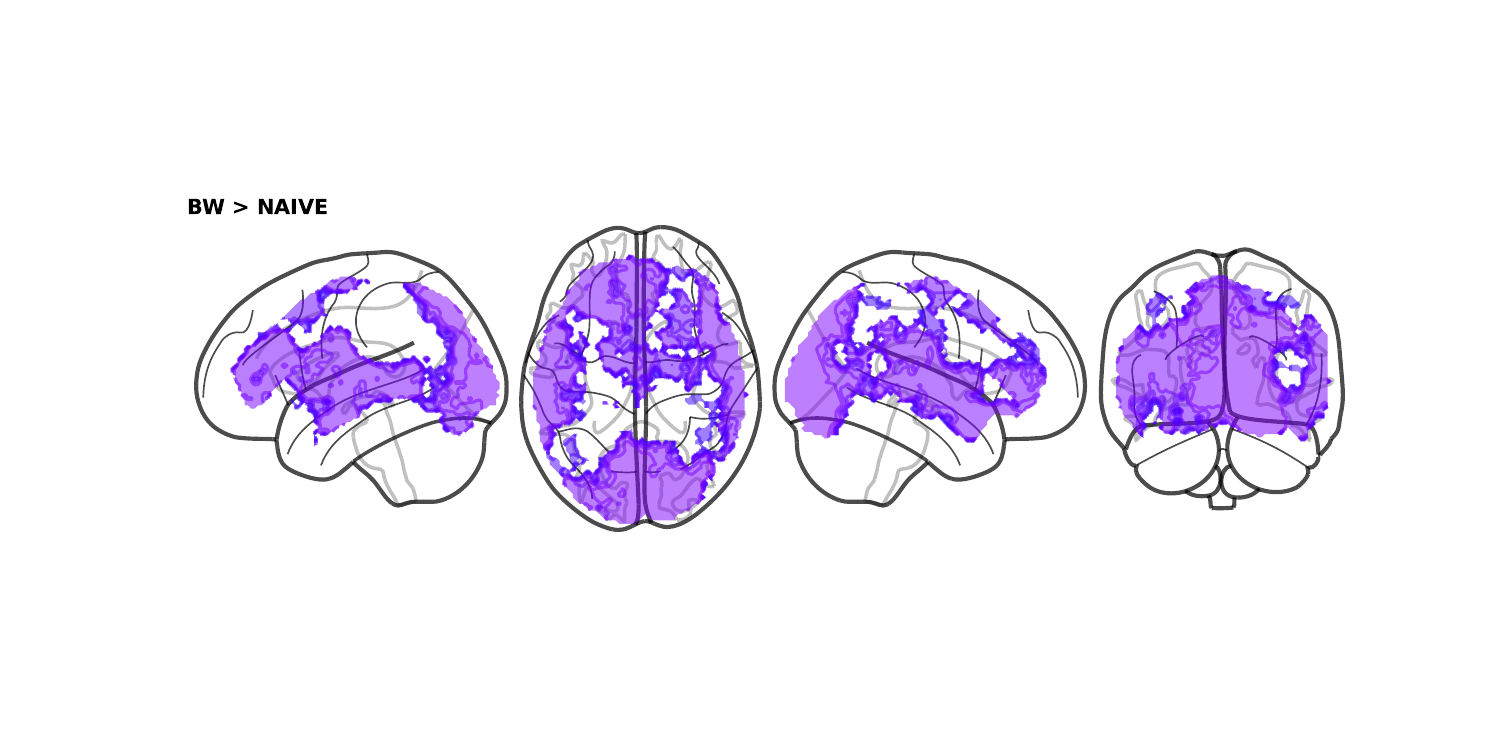}
    \caption{Brain regions where BW decoding is significantly better than the naïve approach ($p<0.05$, whole-brain corrected).}
    \label{fig:projected_sign_BW}
\end{figure}

As illustrated in Supplementary Figure \ref{fig:projected_acc_BW}, the cross-subject average accuracy achieved using BW significantly exceeded chance levels across a broader range of brain regions compared to RTLC. Notably, with BW, our DA-enhanced searchlight successfully classifies imagery of Living versus Non-Living items in regions that extend into the frontal lobe. However, in terms of symmetry between the left and right hemispheres, the accuracy map for BW exhibits less balance compared to RTLC. This asymmetry is likely attributable to cross-subject averaging, as indicated by the comparison between Figure \ref{fig:subj_var} and Supplementary Figure \ref{fig:subj_var_BW}. Specifically, BW demonstrates poor decoding performance for subjects 9, 15, and 17, whereas RTLC achieves successful decoding for these individuals.

\begin{figure}[h]
    \centering
    \includegraphics[width=\textwidth]{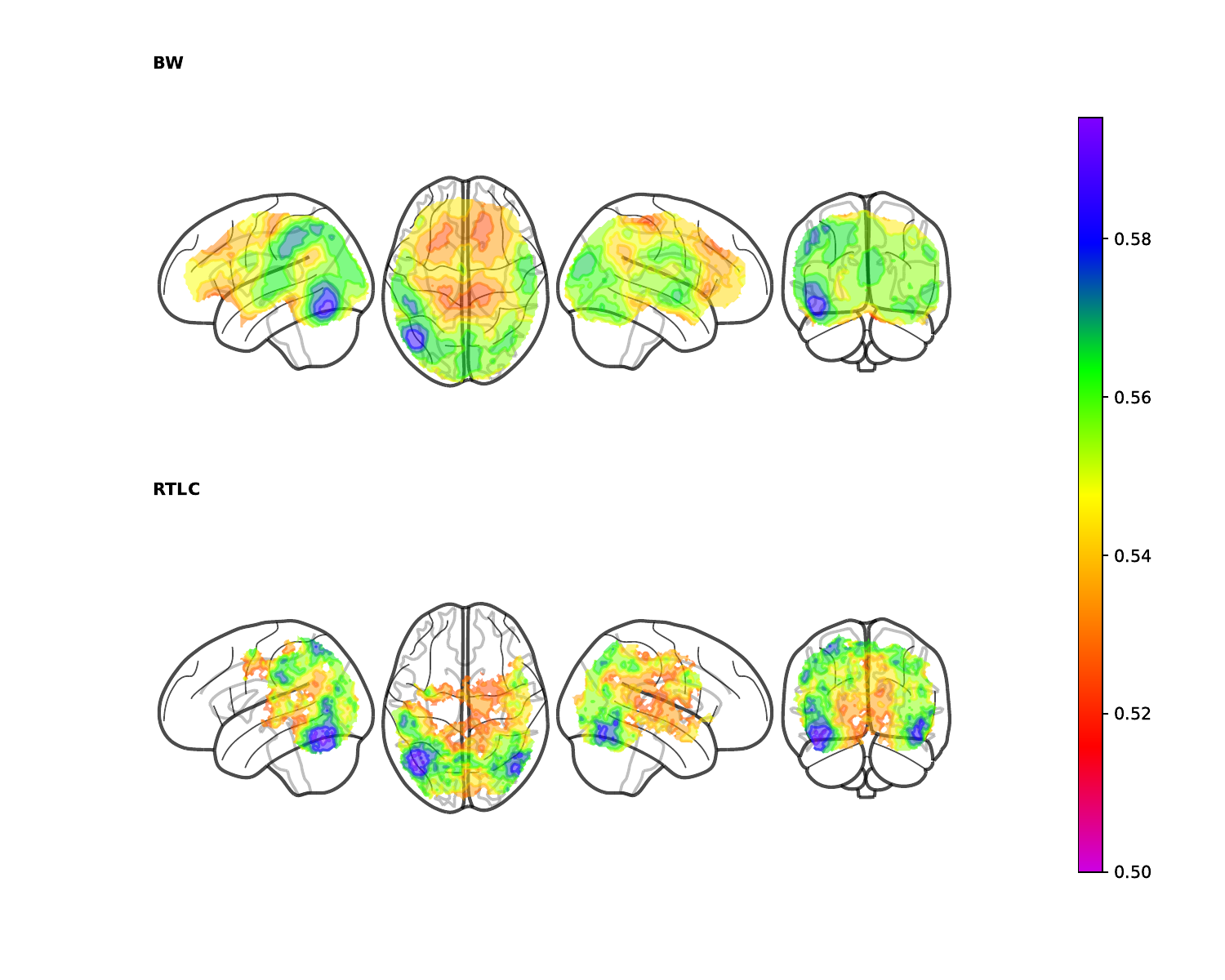}
    \caption{Between-subject average of the balanced accuracy of RTLC and BW in the regions that obtained statistically above chance decoding with each corresponding DA technique.}
    \label{fig:projected_acc_BW}
\end{figure}

\begin{figure}[h]
    \centering
    \includegraphics[width=\textwidth]{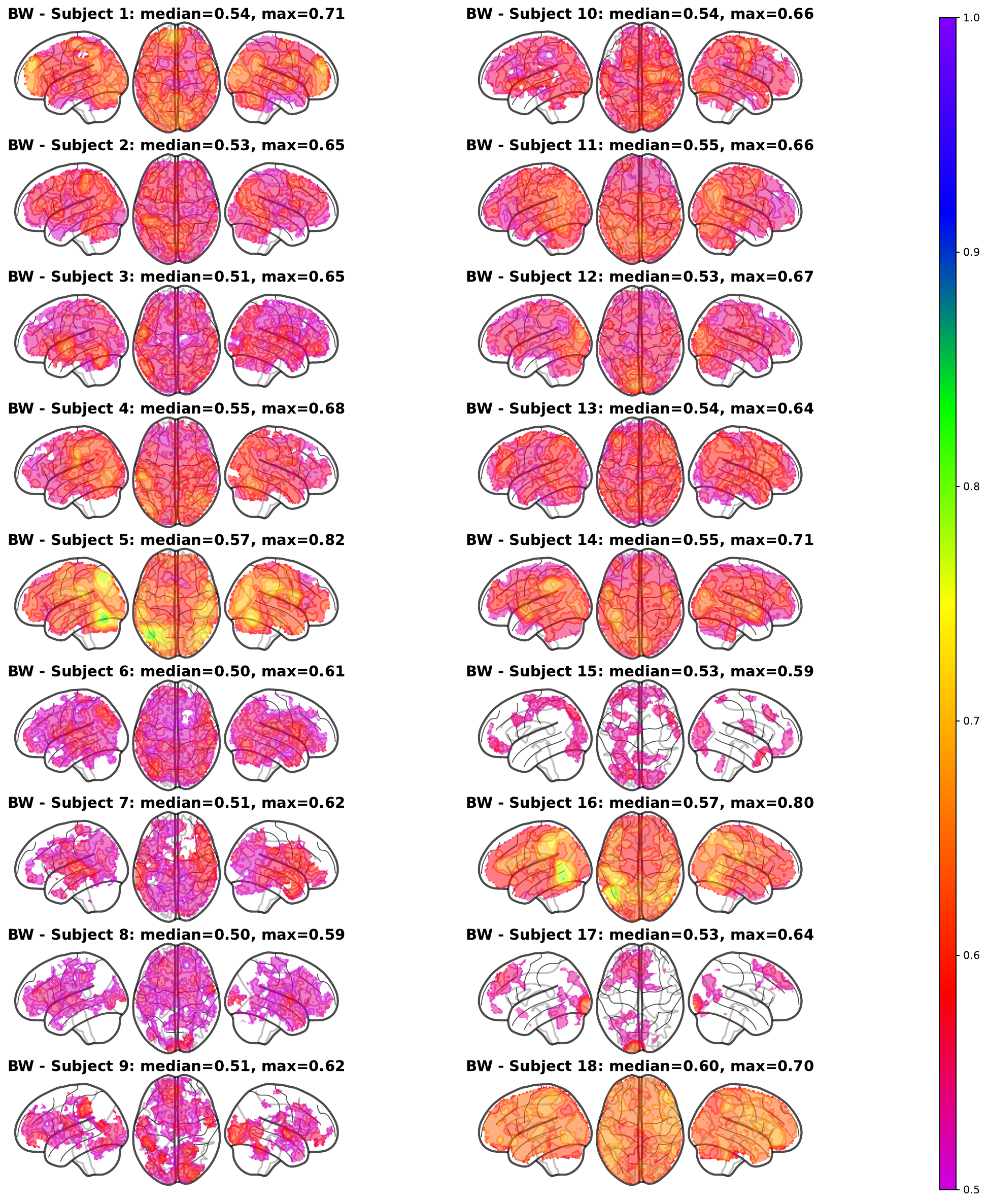}
    \caption{Inter-subject variability of the DA-enhanced searchlight accuracy. For each subject, the colour of each voxel represents the average BW accuracy over the 100 repetitions of the experiment in the regions where such accuracy is statistically above 0.5.}
    \label{fig:subj_var_BW}
\end{figure}

\FloatBarrier

\section{Brief description of each DA method used}
\label{sec:appendix}
\begin{itemize}
    \item \verb|BalancedWeighting| \cite{wu2004improving}: A supervised, instance-based method that fits the base estimator $h$ source and target labelled data according to the modified loss $(1-\gamma)\mathcal{L}(h(X_s),y_s)+\gamma \mathcal{L}(h(X_t),y_t)$, where the parameter $\gamma$ controls the ratio between fitting the source data and adapting to the target data, with higher values indicating more adaptation. In our case, $\mathcal{L}$ is the binary cross-entropy and $\gamma=0.5$. \verb|BalancedWeighting| can be used with any supervised estimator. 
    \item \verb|RegularTransferLC| (Regular Transfer for Linear Classification - RTLC) \cite{chelba2006adaptation}: A supervised, parameter-based method that starts from a source-only parametric linear classifier and translates the classification problem in the target domain into a linear regression task, minimizing the Mean Squared Error with a penalty on the Euclidean distance of the $h_{DA}$ parameters and the source-only model parameters. In short, given a  source-only linear model with parameters $\beta_s$, it constructs a linear model for the target domain with parameters $\beta_t = arg \underset{\beta}{min} || X_t\beta-y_t||^2+\lambda||\beta-\beta_s||$. Higher $\lambda$ indicates a stronger fit to the available target instances, while lower values involve a higher use of the source domain knowledge.
    \item \verb|KMM| (Kernel Mean Matching) \cite{huang2006correcting}: An unsupervised, instance-based sample bias correction approach that minimizes the Maximum Mean Discrepancy (MMD) between source and target domains by solving a quadratic optimization problem such that the means of $X_s$ and $X_t$ in a reproducing kernel Hilbert space (RKHS) are closer than a predefined threshold. We used the default Radial Basis Function kernel.
    \item \verb|TrAdaBoost| (Transfer AdaBoost for Classification) \cite{dai2007boosting}: A supervised, instance-based iterative reweighing approach. Firstly, an initial estimator is fitted on the concatenation of labelled source and target data. Then, the samples are reweighed based on the performance of said estimator, decreasing the weight of the wrongly classified source instances under the hypothesis that those instances are more dissimilar to the ones coming from the target domain.
    \item \verb|FA| \cite{daume2009frustratingly}: A supervised, feature-based pre-processing step leveraging feature augmentation, which aims to separate features into (a) those specific to the source domain, (b) those specific to the target domain and (c) those common to both, without altering the model fitting stage, so it can be applied to any supervised estimator. 
    \item \verb|PRED| \cite{daume2009frustratingly}: Another feature-augmentation DA technique, mixing the idea in \verb|FA| with the notion of model stacking. First, the base estimator is trained solely on source data, predicting target values on the available target data. These predictions are concatenated with $X_t$ as a new feature, and a new instance of the estimator is trained on the resulting data-set to obtain the final $h_{DA}$.
    
    \item \verb|ULSIF| (Unconstrained Least-Squares Importance Fitting) \cite{kanamori2009least}: An unsupervised, instance-based method to correct the difference between input distributions of source and target domains as measured by their relative Pearson divergences by reweighing the source instances. This entails solving a quadratic optimization problem, which has a closed-form solution involving the inverse of a matrix $H$ of size equal to the number of samples in $X_s$. This matrix contains the element-wise products of a kernel transformation applied to $X_s$. $H^{-1}$ is then multiplied with the kernel transformation of $X_t$ to give the weights of the source samples. We used an RBF kernel.
    \item \verb|NearestNeighborsWeighting| (NNW) \cite{loog2012nearest}: An unsupervised, instance-based approach that computes, for each source domain sample, the number of neighbours inside a certain radius in the target data set. Source instances are reweighted proportionately.
    \item \verb|RULSIF| (Relative Unconstrained Least-Squares Importance Fitting) \cite{yamada2013relative}: A very similar approach to ULSIF, but allowing the target instances to influence the matrix $H$ via an additive contribution of the element-wise products of their kernel transformation.  
     \item \verb|SA| (Subspace Alignment) \cite{fernando2013unsupervised}: An unsupervised, feature-based method that linearly aligns the source domain to the target domain in a reduced PCA subspace of a certain pre-specified dimension.
    \item \verb|IWN| (Importance Weighting Network) \cite{de2022fast}: An unsupervised, instance-based method that reweighs the source domain samples minimizing the MMD between the reweighed source and the target distributions by using a neural network to predict the sample weights, thus reducing the computational burden of KMM. 
    \item \verb|FineTuning| (FT) \cite{finetuning}: A feature-based, supervised DA method originally developed for Convolutional Neural Networks, that reuses the internal layers of a network trained in the source domain and furthers training of the external layers using a limited amount of target domain data.
    \item \verb|DeepCORAL| (Deep CORrelation ALignment - DCORAL) \cite{deepcoral}: Correlation Alignment is a feature-based and unsupervised DA method that aligns the second-order statistics of the source and target distributions using a linear transformation. Similarly, Deep CORAL is a nonlinear extension that aligns correlations of layer activations in deep neural networks.
    \item \verb|DANN| (Discriminative Adversarial Neural Network) \cite{DANN}: DANN is a feature-based DA method, which strives to find a new representation of the input features in which source and target data are indistinguishable by any discriminator network. This new representation is given by an encoder which is learned along the discriminator using a reversal gradient layer. In parallel, a task network is learned on the encoded space. 
    \item \verb|MCD| (Maximum Classifier Discrepancy) \cite{MCD}: MCD is a feature-based, unsupervised DA method that looks for a new representation of the input features which minimizes the discrepancy between the source and target domains. The discrepancy is computed through adversarial training of three networks: An encoder and two classifiers, each of which learn the task on the source and the target domains. A reversal layer is placed between the encoder and the two classifiers to perform adversarial training.
    
\end{itemize}

\end{appendices}


\bibliography{full_bibliography}

\end{document}